\pdfoutput=1
\documentclass[11pt]{article}

\usepackage[final]{acl}

\usepackage{times}
\usepackage{latexsym}
\usepackage{amsmath}
\usepackage{amssymb}
\usepackage{amsthm}
\usepackage{graphicx}

\usepackage[english, vietnamese]{babel}

\usepackage[T1]{fontenc}
\usepackage[utf8]{inputenc}

\usepackage{microtype}

\usepackage{inconsolata}

\usepackage{graphicx}

\renewcommand\abstractname{summary}
\title{Using Large Language Models for education managements in Vietnamese with low resources}

\author{Duc Do Minh \\
  UET-VNU \\
Vietnam National University \\
Hanoi, Vietnam \\
  \texttt{ducdm103.work@gmail.com} \\ \And
  \hspace{1cm}Vinh Nguyen Van \\
  \hspace{1cm} UET-VNU \\
\hspace{1cm}Vietnam National University \\
\hspace{1cm}Hanoi, Vietnam \\
  \hspace{1cm} \texttt{vinhnv@vnu.edu.vn} \\ \And
  \hspace{1cm} Thang Dam Cong \\
  \hspace{1cm} Bac Ninh Teacher Training College\\
\hspace{1cm} Bac Ninh, Vietnam \\
  \hspace{1cm} \texttt{damcongthang@cdspbacninh.edu.vn} \\
  }

\begin{document}
\maketitle
% \addcontentsline{toc}{abstract}{Abstract}
\renewcommand{\abstractname}{Abstract}
\renewcommand{\figurename}{Figure}
\renewcommand{\tablename}{Table}

\begin{abstract}

% Large language models, such as GPT4o, Gemini 1.5, Claude 3.5 Sonnet, Lama3 ..., have shown great success in numerous tasks for NLP since the release of ChatGPT in 2022. However, fine-tuning and using LLMs are very computationally expensive. However, fine-tuning and using LLMs are very computationally expensive. In this paper, we present a simple and effective framework for applying large language models (LLMs) to educational management tasks. Experiments with dataset , collected by student education documents in Hanoi VNU, show that our methods are better than previous methods.

Large language models (LLMs), such as GPT-4, Gemini 1.5, Claude 3.5 Sonnet, and Llama3, have demonstrated significant advancements in various NLP tasks since the release of ChatGPT in 2022. Despite their success, fine-tuning and deploying LLMs remain computationally expensive, especially in resource-constrained environments. In this paper, we proposed VietEduFrame, a framework specifically designed to apply LLMs to educational management tasks in Vietnamese institutions. Our key contribution includes the development of a tailored dataset, derived from student education documents at Hanoi VNU, which addresses the unique challenges faced by educational systems with limited resources. Through extensive experiments, we show that our approach outperforms existing methods in terms of accuracy and efficiency, offering a promising solution for improving educational management in under-resourced environments. While our framework leverages synthetic data to supplement real-world examples, we discuss potential limitations regarding broader applicability and robustness in future implementations.

\end{abstract}

\section{Introduction}
Most current tasks in Natural Language Processing (NLP) are dominated by large language models (LLMs) such as GPT4 and Gemini 1.5, which have set new benchmarks for performance. These models excel in a wide range of applications, demonstrating superior capabilities in understanding and generating human language.

In recent years, artificial intelligence (AI) and machine learning (ML) for education have received a great deal of interest and have been applied in various educational scenarios \cite{Chen2020ArtificialII}, \cite{Xia2022SystematicLR}, \cite{Latif2023ArtificialGI}, \cite{Denny2023ComputingEI}, \cite{Li2024BringingGA}. Educational data mining methods have been widely adopted in different aspects such as cognitive diagnosis \cite{Batool2022EducationalDM}, knowledge tracking \cite{Koedinger2015DataMA}, and specifically question answering \cite{Lende2016QuestionAS}, \cite{Thiruvanantharajah2021AutomatedQA}, \cite{Bhowmick2023AutomatingQG}.

Large language models (LLMs) have emerged as a powerful paradigm across different areas \cite{Chen2023ExploringTP}, \cite{Fan2023RecommenderSI}, \cite{Jin2024PositionWC}, \cite{Zeng2023LargeLM}, and have achieved state-of-the-art performances in multiple educational scenarios \cite{Kasneci2023ChatGPTFG}, \cite{Li2023AdaptingLL}, \cite{Yan2023PracticalAE}. Existing work has found that LLMs can achieve student-level performance on standardized tests in a variety of subjects, including mathematics, physics, and computer science, on both multiple-choice and free-response problems. A recent study \cite{Susnjak2022ChatGPTTE} reveals that ChatGPT is capable of generating logically consistent answers across disciplines, balancing both depth and breadth. Another quantitative analysis \cite{Malinka2023OnTE} shows that students using ChatGPT (by keeping or refining the results from LLMs as their own answers) perform better than average students in some courses in the field of computer security.

Despite the global advancements, there remains a significant gap in the application of these technologies within the context of Vietnamese education, particularly in educational management. My research is among the first in Vietnam to explore these applications broadly in education and specifically in educational management. Due to the limitations of resources and data within Vietnamese institutions, this area has not yet received adequate attention. This scarcity of local studies and resources has driven us to undertake this research, aiming to bridge the gap and contribute to the growing body of knowledge in this critical field.

In this study, our main contributions can be summarized as follows: 

\begin{itemize}
    \item \textbf{Framework Proposal:} We propose a simple yet highly effective framework for applying large language models (LLMs) to educational management tasks. This framework is designed to be easily implementable and adaptable within the constraints of Vietnamese educational institutions. To the best of our knowledge, this first study focuses applying LLMs for education in Vietnamese.
    
    \item \textbf{New Dataset:} We introduce a new dataset specifically tailored for educational management in Vietnam. This dataset addresses the unique challenges and characteristics of the Vietnamese educational context, providing a valuable resource for future research and development.
    
    \item \textbf{Model Development with Limited Resources:} We successfully develop and deploy a model using the limited computational resources available at our institution. This demonstrates the feasibility of implementing advanced AI solutions in resource-constrained environments and provides a blueprint for similar institutions. 
\end{itemize}

\section{Related work}
\subsection{Large language models in for study assisting}
Providing students with timely learning support has been widely recognized as crucial in improving student engagement and learning efficiency during their independent studies \cite{Dewhurst2000IndependentSL}. Due to the limitation of prior algorithms in generating fixed-form responses, many of the existing study-assisting approaches face poor generalization challenges while being implemented in real-world scenarios \cite{Knig2022CriticalSF}. Fortunately, the appearance of LLMs brings revolutionary changes to this field. Using finetuned LLMs \cite{Ouyang2022TrainingLM} to generate human-like responses, recent studies in LLM-based educational support have demonstrated promising results.

Contributing to the large-scale parameter size of LLMs and the enormous sized and diverse web corpus used during the pre-training phase, LLMs have been proven to be a powerful question zero-shot solver to questions spread from a wide spread of subjects, including math \cite{Wu2023AnES} \cite{Yuan2023HowWD}, law \cite{Bommarito2022GPTTT} \cite{Cui2023ChatlawAM}, medicine \cite{Lievin2022CanLL} \cite{Thirunavukarasu2023LargeLM}, finance \cite{Wu2023BloombergGPTAL} \cite{Yang2023FinGPTOF}, programming \cite{Kazemitabaar2023HowNU} \cite{avelka2023ThrilledBY}, language understands\cite{Zhang2023M3ExamAM}. In addition, to further improve LLM’s problem-solving performance while facing complicated questions, various studies have been actively proposed. For example, \cite{Wei2022ChainOT} proposes the Chain-of-Thought (CoT) prompting method, which guides LLMs to solve a challenging problem by decomposing it into simpler sequential steps. Other works exploit the strong in-context learning ability of LLMs and propose advanced few-shot demonstration-selection algorithms to improve LLM’s problem-solving performance to general questions. \cite{Chen2022ProgramOT} and \cite{Gao2022PALPL} leverage external programming tools to avoid calculation errors introduced during the textual problem-solving process of raw LLMs. \cite{Wu2023AutoGenEN} regard chat-optimized LLMs as powerful agents and design a multi-agent conversation to solve those complicated questions through a collaborative process.

\subsection{Education toolkit}
Utilizing a chatbot powered by a Large Language Model (LLM) as an educational tool presents numerous benefits and opportunities. LLM chatbots can tailor their responses to meet the unique needs of each learner, offering personalized feedback and assistance. This ability to customize can cater to various learning styles, speeds, and preferences. They are available 24/7, making learning accessible at any time and from any place, which is especially advantageous for learners in different time zones or with diverse schedules. The interactive features of chatbots can make learning more engaging and enjoyable. They can mimic conversations, set up interactive learning scenarios, and give immediate feedback, which can be more effective than passive learning approaches. Chatbots can manage thousands of inquiries at once, providing a scalable solution for educational institutions to support a large number of students without needing more teaching staff. They can also automate repetitive teaching tasks, such as grading quizzes or offering basic feedback, enabling educators to concentrate on more complex and creative teaching duties. Notable examples of such chatbots include ChatGPT, Bing Chat, Google Bard, Perplexity, and Pi Pi.ai.

\subsection{Textbook question answering}
Textbook Question Answering (TQA) is a task that requires a system to comprehensively understand the multi-modal information from the textbook curriculum, spreading across text documents, images, and diagrams. The major challenge of textbook question answering is to comprehend the multimodal domain-specific contexts as well as the questions, and then identify the key information to the questions.

\textbf{Datasets} \cite{Kembhavi2017AreYS} presented the TQA dataset, designed to assess a system that integrates multi-modal contexts and a wide range of scientific topics. Comparable datasets, such as AI2D \cite{Kembhavi2016ADI}, DVQA \cite{Kafle2018DVQAUD}, and VLQA \cite{Sampat2020VisuoLingusticQA}, have been developed to facilitate research in multi-modal reasoning within the scientific domain. Nonetheless, these datasets lack annotated explanations for answers in the form of supporting facts. SCIENCEQA \cite{Lu2022LearnTE} is a comprehensive textbook question-answering dataset that includes annotated lectures and explanations. This dataset is derived from elementary and high school science curricula, covering a variety of science topics such as natural science, social science, and language science. Recently, the TheoremQA dataset has been released, which includes textbook questions at the university level \cite{Chen2023TheoremQAAT}. Beyond the scientific domain, there are datasets focused on the medical field. MEDQA \cite{Jin2020WhatDD} and MedMCQA \cite{Pal2022MedMCQAA} are two medical question-answering datasets that encompass a broad range of healthcare topics, derived from both real-world scenarios and simulated exams.

\textbf{Methods}. From a technical perspective, textbook question answering is inherently similar to visual question answering (VQA) \cite{Dosovitskiy2020AnII}, \cite{Gao2018DynamicFW}, \cite{Gao2022TransformRetrieveGenerateNL}. Traditional VQA approaches use RNNs to encode the question and CNNs to encode the image \cite{Agrawal2015VQAVQ}, \cite{Malinowski2015AskYN}. The multi-modal information is then fused to understand the questions. Additionally, other methods that utilize spatial attention \cite{Lu2016HierarchicalQC}, \cite{Noh2016TrainingRA}, \cite{Xu2015ShowAA}, \cite{Yang2015StackedAN}, compositional strategies \cite{Andreas2016LearningTC}, and bilinear pooling schemes \cite{Fukui2016MultimodalCB}, \cite{Liu2022APM} have been proposed to enhance VQA performance.

While VQA and textbook question answering share significant similarities, textbook question answering requires domain-specific knowledge for the accompanying context and innovative integration of diagrams and tables. To address this gap, \cite{Ram2021FewShotQA} proposed a pre-training schema tailored for question answering. Specifically, their method improves performance in textbook question answering by masking recurring span selections and selecting the correct span in the passage, even when only a hundred examples are available in specific domains. An adversarial training framework is also adapted for domain generalization \cite{Lee2019DomainagnosticQW}, enabling question-answering models to learn domain-invariant features. \cite{Xu2022FromCT} introduced a novel Pre-trained Machine Reader as an enhancement of pre-trained Masked Language Models (MLMs), which addresses the discrepancy between model pre-training and downstream fine-tuning for specific domain MLMs. To comprehend diagrams and tables, graph-based parsing methods have been developed to extract concepts from diagrams \cite{Kembhavi2016ADI} by converting a diagram into a diagram parse graph. Optical Character Recognition (OCR) is employed to identify chart-specific answers from the charts, which are then aligned with the questions \cite{Poco2017ReverseEngineeringVR}, \cite{Kafle2018DVQAUD}.

Our research is different from previous works in some significant ways: 
\begin{itemize}
    \item \textbf{First,} we have developed a simple yet effective framework for the textbook question-answering problem. This framework has proven to be both efficient and robust, delivering high performance within a short development cycle.
    \item \textbf{Second,} leveraging this framework, we have created a dedicated dataset specifically tailored for the training management process at the Vietnam National University of Hanoi. This dataset is instrumental in enhancing the quality and effectiveness of training management, marking a substantial contribution to the educational resources available for Vietnamese institutions.
\end{itemize}

\section{Dataset}
The use of data in the field of educational management presents several significant challenges, particularly when developing a question-answering system for the Vietnamese language. These challenges include:

\begin{itemize}
    \item \textbf{Institutional Variability:} Each educational institution must comply with the regulations set forth by the Ministry of Education. However, beyond these mandatory guidelines, institutions often have additional rules and policies specific to their own organization or the larger entity they are affiliated with. This variability can lead to inconsistencies in data structure, terminology, and reporting practices, complicating the task of creating a unified dataset.

    \item \textbf{Data Standardization:} Due to the diverse regulatory requirements and internal policies across different institutions, standardizing data becomes a complex process. Ensuring consistency and compatibility of data from various sources is essential for effective analysis and model training but is difficult to achieve given the heterogeneity of the data.

    \item \textbf{Data Availability and Quality:} As one of the first studies addressing the question-answering problem in the Vietnamese language within the educational management domain, there is a scarcity of readily available datasets. Existing datasets in other languages or educational contexts may not be directly applicable due to linguistic and contextual differences. Therefore, sourcing high-quality data externally is challenging, necessitating the creation of a new dataset from scratch.

    \item \textbf{Data Collection and Annotation:} Building a new dataset requires significant effort in data collection and annotation. This process involves gathering data from various educational institutions, ensuring its accuracy and relevance, and annotating it to create a structured dataset suitable for training machine learning models. The annotation process, in particular, is time-consuming and demands a deep understanding of the educational domain.

\end{itemize}

Addressing these challenges is crucial for the success of our research. By acknowledging and systematically tackling these issues, we aim to build a robust and reliable dataset that will facilitate the development of effective AI solutions for educational management in Vietnam.

\subsection{Building data}
In this subsection, I describe the process of constructing a dataset from the "Regulations on Student Affairs of Vietnam National University" (VNU) to train a model for question-answering tasks. By using prompts, we generate data points that each consist of a "context," "question," and "answer." This structured approach ensures comprehensive coverage of the regulations and facilitates the creation of a robust dataset for training.  The process consists of five critical steps: data preprocessing, data analysis and prompt design, data generation using prompts and LLMs, and data quality evaluation.

The first step \textbf{data preprocessing} involves preparing the raw data for subsequent analysis and prompt generation. This includes:
\begin{itemize}
    \item\textbf{Data Cleaning:} Removing any irrelevant information, and duplicates, and ensuring consistency in formatting.
    \item \textbf{Text Segmentation:} Breaking down the regulations into manageable sections that can be used as context for generating questions and answers.
    \item \textbf{Whitespace and Extraneous Character Removal:} Removing unnecessary spaces and characters to ensure clean text.
    \item \textbf{Spell Checking:} Correcting any spelling errors in the text.
    \item \textbf{Math Formula Conversion:} Converting mathematical formulas into KATEX format for consistent representation.
\end{itemize}

After preprocessing the data, the next step is to \textbf{analyze the content and design effective prompts}. This involves:
\begin{itemize}
    \item \textbf{Content Analysis:} Identifying key themes, rules, and guidelines within the regulations.
    \item \textbf{Prompt Crafting:} Developing specific prompts that will be used to generate questions and answers. Each prompt focuses on different aspects of the regulations, ensuring comprehensive coverage.
    \item \textbf{Using technique prompting} Chain of Thought, Self-Consistency Chain of Thought, and Tree of Thought: Employing advanced prompting techniques to enhance the generation process.
\end{itemize}

The next steps are \textbf{data generation using prompts and LLMs, and data quality evaluation}. To generate the desired dataset, we utilized prompts that were meticulously designed in the previous phase. These prompts were fed into large language models (LLMs) such as GPT-3,5 turbo, which then generated a comprehensive set of synthetic data. The generation process was systematic and aimed to produce data that closely aligned with our research objectives and covered the necessary range of scenarios.

The quality of the generated data was evaluated using both automated metrics and human assessment. Specifically, we employed ROUGE and BLEU scores to quantify the relevance and coherence of the generated text. These metrics provided an objective measure of how well the generated data matched the expected output in terms of n-gram overlap and sequence similarity.

In addition to automated metrics, human evaluators conducted a qualitative review of the generated data. These domain experts assessed the data for relevance, coherence, and diversity, ensuring that the synthetic data met the high standards required for our study. This dual approach of combining quantitative scores with qualitative human judgment ensured a robust evaluation of the generated dataset, confirming its suitability for subsequent analyses and experiments.

Here is an example of the dataset

\begin{figure}[!hbt]
    \centering
    \caption{The examples of Question Answering in the education domain}
    \includegraphics[width=\linewidth]{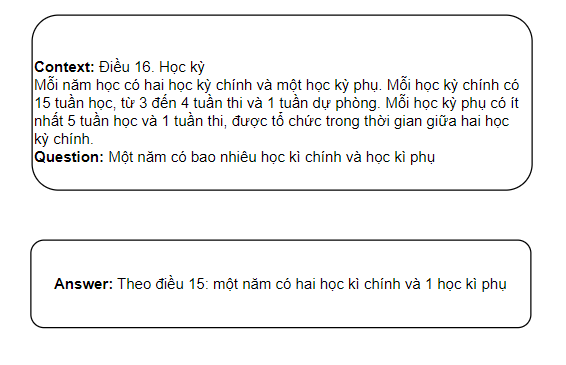}
    \label{fig:example-label}
\end{figure}

\section{Methodology}
\begin{figure*}[!hbt]
  \centering
  \caption{Overview of our framework}
  \includegraphics[width=0.9\linewidth]{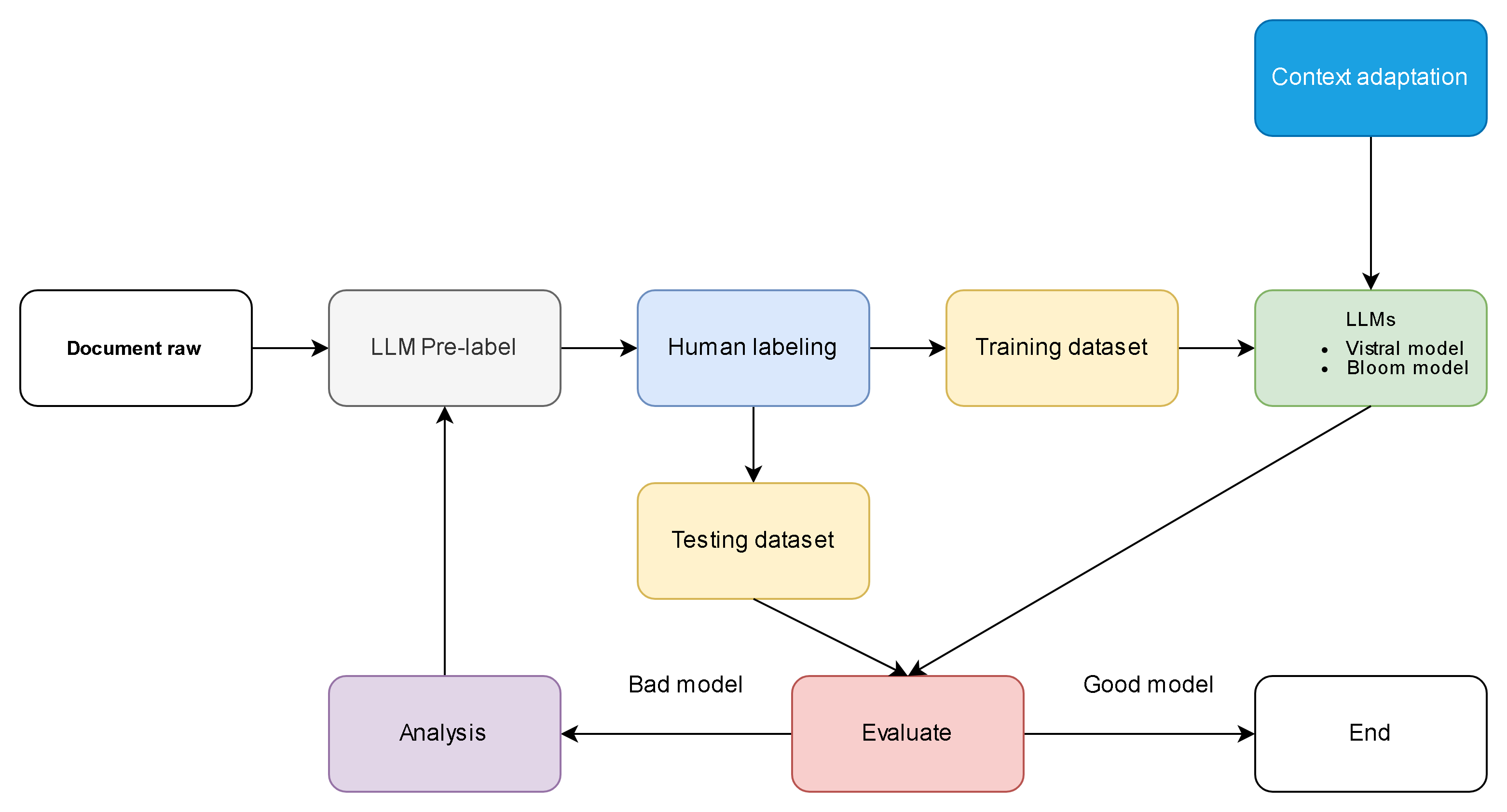}
  
    \label{fig:framework}
\end{figure*}

In this section, we detail the methodology employed to address the question-answering problem within the domain of university educational management in \ref{fig:framework}. Our approach encompasses several key stages: leveraging a Large Language Model (LLM) for initial data pre-labeling, human labeling for data refinement, training the model, evaluating its performance, and conducting a thorough analysis of the results. Each step in this pipeline is meticulously designed to ensure accuracy and effectiveness, tailored to the specific needs and constraints of the educational context in Vietnam.

We will systematically describe each stage of our methodology as follows:
\begin{itemize}
    \item \textbf{Large Language Model (LLM):}An overview of the LLM utilized in our study, highlighting its features and advantages in handling natural language processing tasks.
    \item \textbf{Pre-labeling:} A description of the pre-labeling process using the LLM to provide initial annotations for the dataset, which sets a foundation for further refinement.
    \item \textbf{Human Labeling:} An explanation of the human labeling process, emphasizing its role in ensuring high-quality data by correcting and improving the initial LLM-generated labels.
    \item \textbf{Training:} Details on the training phase, including the algorithms and techniques applied to build a robust question-answering model.
    \item \textbf{Evaluation:} Presentation of the evaluation methods and criteria used to assess the model's performance, ensuring it meets the desired standards of accuracy and reliability.
    \item \textbf{Analysis:}A comprehensive analysis of the results obtained from the evaluation, providing insights into the model's strengths and areas for improvement.
\end{itemize} 
 
\subsection{Pre labeling and human labeling}
With two steps using LLMs pre-labeling và human labeling, I illustrated in section 3 building data.
\subsection{Training and context adaptation}
In this subsection, we describe the training process and context adaptation techniques employed to enhance the question-answering capabilities of our model, particularly tailored to university educational management.

\subsubsection{Training}

\textbf{Model Vistral}

Vistral \cite{Vo2024ViMistralXBA} is a deep learning model that uses many transformer decoder layers to generate coherent and natural language text. The model was pre-trained on a large corpus of text data using an unsupervised learning approach, which enabled it to learn the statistical patterns and structures of natural language. Vistral has been widely used for various NLP tasks such as language translation, question-answering, text summarization, and even creative writing. As of now April 2024, the Vistral model is the highest-scoring public model on the VMLU leaderboard. Vistral model is an innovative Large Language Model designed expressly for the Vietnamese language.

\begin{itemize}
    \item Rolling Buffer Cache
    \item Sliding-Window Attention
    \item Pre-fill and Chunking
\end{itemize}

\textbf{Sliding Window Attention} utilizes the multiple layers of a transformer to access information beyond a defined window size W. In this method, the hidden state at position $i$ in layer $k$, denoted as $h_i$, attends to all hidden states in the preceding layer within the range from $i - \textbf{W}$ to $i$. This process allows $h_i$ to recursively access tokens from the input layer at a distance of up to $\textbf{W} \times k$ tokens.

\textbf{Rolling Buffer Cache}. By having a fixed attention span, we can manage our cache size with a rolling buffer cache. This cache has a set size of $\textbf{W}$, and the keys and values for timestep $i$ are saved in the position $i mod \textbf{W}$ of the cache. Consequently, when position $i$ exceeds $\textbf{W}$, the older values in the cache are overwritten, preventing the cache size from growing indefinitely.

\textbf{Pre-fill and Chunking}. When generating a sequence, tokens must be predicted one at a time, as each token depends on the previous ones. However, since the prompt is known beforehand, we can pre-fill the (k, v) cache with the prompt. If the prompt is very large, it can be divided into smaller chunks, and the cache can be pre-filled with these chunks. The window size can be used as the chunk size. For each chunk, it is necessary to compute the attention over both the cache and the chunk.

\textbf{Model Bloom}
BLOOM is a powerful autoregressive Large Language Model (LLM) designed to extend text from a given prompt, utilizing extensive computational resources on massive text datasets. This capability allows it to produce fluent text in 46 different languages and 13 programming languages, making it almost indistinguishable from human-written content. Additionally, BLOOM can be directed to undertake text-related tasks it wasn't specifically trained for by framing them as text-generation problems.

Modeling Details
Several key innovations were incorporated into the BLOOM model to enhance its performance and stability:

\textbf{ALiBi Positional Embeddings}: The model employs ALiBi (Attention Linear Bias) positional embeddings instead of traditional positional embeddings. ALiBi attenuates attention scores based on the distance between keys and queries, which results in smoother training dynamics and improved performance.

\textbf{Embedding LayerNorm}: An additional layer normalization step is applied immediately after the embedding layer. This modification was implemented to improve training stability, especially considering the use of bfloat16 precision in the final training phase, which offers more stability than float16.

\textbf{Low rank Adaptation}

For a given pretrained weight matrix $W_{0} \in \mathbb{R}^{d \times k}$, LoRA introduces two trainable weight matrices, $W_{up} \in \mathbb{R}^{d \times r}$ and $W_{down} \in \mathbb{R}^{r \times k}$ where the rank $r \ll \min(d, k)$, operating in parallel to $W_{0}$. Let represent the input. Under normal conditions, the output through $W_{0}$ is $h_{out} = W_{0} h_{in}$. Instead, LoRA modifies this output by introducing an incremental update $\Delta W$ that encapsulates task-specific knowledge:

\begin{equation}
    h_{\text {out }}=W_0 h_{\text {in }}+\frac{\alpha}{r} \Delta W h_{\text {in }}=W_0 h_{\text {in }}+\frac{\alpha}{r} W_{\text {up }} W_{\text {down }} h_{\text {in }}
\end{equation}

where $\alpha$ denotes a scaling factor. At the onset of training, $W_{down}$ is initialized using a random Gaussian distribution, while $W_{up}$ is initialized to zero, ensuring that $\Delta W$ initially holds a value of zero. LoRA is straightforward to implement and has been evaluated on models with up to 175 billion parameters. In this research, I use this method for the model Bloom and Vistral-7B. Once fine-tuning is complete, LoRA’s adaptive weights seamlessly integrate with the pre-trained backbone weights. This integration ensures that LoRA maintains the model’s efficiency, adding no extra burden during inference. The number of parameters training is reduced $dk/(d + k) / r$ times.

\subsubsection{Context Adaptation}
Context adaptation is crucial for activating the model's question-answering capabilities. We enhance the training data by incorporating detailed instructions and contextual cues that guide the model in understanding and generating accurate responses to educational queries.

By adding specific instructions, we provide the model with explicit examples of how to approach different types of questions within the educational domain. These instructions act as triggers, enabling the model to apply its learned knowledge effectively and respond accurately to complex queries.

Our training and context adaptation approach ensures that the models are not only finely tuned to our dataset but also contextually aware, enhancing their ability to provide precise and relevant answers in the context of university educational management. The combination of dual-model training and LoRA, along with detailed context adaptation, significantly boosts the model's performance and usability in real-world applications.

\subsection{Evaluate}
\textbf{Exact Match (EM)}: For each question-answer pair, if the characters of the MRC system's predicted answer exactly match the characters of (one of) the gold standard answer(s), EM = 1, otherwise EM = 0. EM is a stringent all-or-nothing metric, with a score of 0 for being off by a single character. When evaluating against a negative question, if the system predicts any textual span as an answer, it automatically obtains a zero score for that question.

\textbf{F1-score}: F1-score is a popular metric for natural language processing and is also used in machine reading comprehension. F1-score is estimated over the individual tokens in the predicted answer against those in the gold standard answers. The F1-score is based on the number of matched tokens between the predicted and gold standard answers.
\begin{equation}
\text{Precision} = \frac{\text{the number of matched tokens}}{\text{the total tokens in the predicted answer}}    
\end{equation}
\begin{equation}
\text{Recall} = \frac{\text{the number of matched tokens}}{\text{the total tokens in the gold standard answer}}    
\end{equation}
\begin{equation}
\text{F1-score} = \frac{2 \times \text{Precision} \times \text{Recall}}{\text{Precision} + \text{Recall}}    
\end{equation}

\section{Result and Experiment}
\subsection{Statistic of dataset}
In this subsection, we present a comprehensive statistical analysis of our dataset, which includes an in-depth survey of the lengths and averages of contexts, questions, and answers. Understanding these metrics is crucial for evaluating the overall quality and characteristics of the data used in our experiments.
\begin{table}[!hbt]
\caption{Statistic of dataset}
\begin{tabular}{|p{0.5in}|p{0.5in}|p{0.5in}|p{0.5in}|}
\hline
      & context length & question length & answer length \\ \hline
count & 985.00      & 985.00       & 985.00     \\ \hline
mean  & 882.48      & 74.03        & 415.60     \\ \hline
std   & 742.12      & 32.59        & 342.66     \\ \hline
min   & 49.00       & 15.00        & 21.00      \\ \hline
25\%  & 324.00      & 54.00        & 166.00     \\ \hline
50\%  & 611.00      & 71.00        & 298.00     \\ \hline
75\%  & 1371.00     & 86.00        & 569.00     \\ \hline
max   & 4446.00     & 289.00       & 2163.00    \\ \hline
\end{tabular}
\end{table}

\begin{figure}[!hbt]
    \centering
    \includegraphics[width=0.95\linewidth]{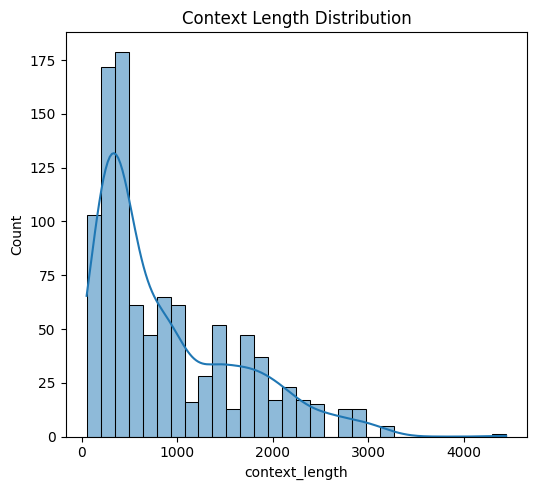}
    \caption{Context Length Distribution}
    \label{fig:context-length-distribution}
\end{figure}

\begin{figure}[!hbt]
    \centering
    \includegraphics[width=0.95\linewidth]{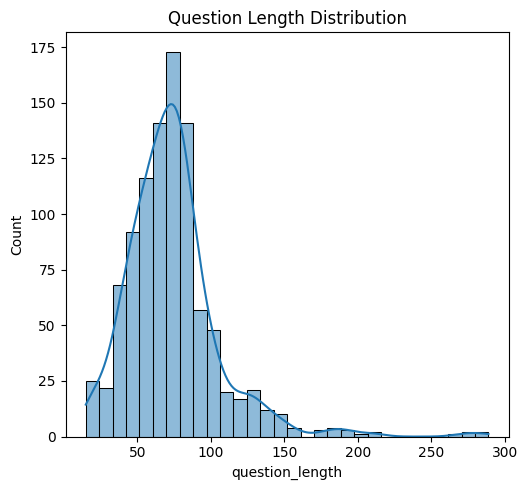}
    \caption{Question Length Distribution}
    \label{fig:question-length-distribution}
\end{figure}

\begin{figure}[!hbt]
    \centering
    \includegraphics[width=0.95\linewidth]{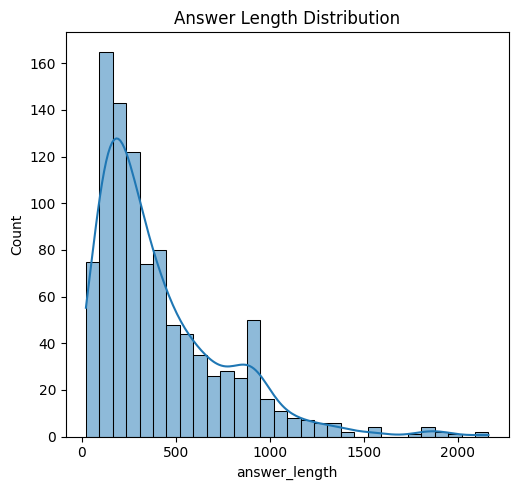}
    \caption{Answer Length Distribution}
    \label{fig:answer-length-distribution}
\end{figure}

\subsection{Data review}
In our study, we categorize the dataset into five distinct levels of question-answering data quality: Very Good, Good, Medium, Bad, and Very Bad. These levels are comprehensively described in Table \ref{table:levels-of-data-quality}

\begin{table}[ht!]
\small\addtolength{\tabcolsep}{-5pt}
\hspace*{-0.3cm}
\caption{Levels of Data Quality in Question Answering}
\centering
\label{table:levels-of-data-quality}

\begin{tabular}{|p{0.7in}|p{2in}|}
\hline
  \textbf{Type} &
  \textbf{Description} \\
\hline
  Very good &
  Answers at this level are completely accurate and directly address the question posed. They exhibit a perfect understanding of the query and provide comprehensive, precise information. The content is well-structured and leaves no room for ambiguity. \\
\hline
Good &
  Answers in this category are mostly accurate and address the main aspects of the question. They may lack some minor details or have slight imprecisions but still provide a reliable and useful response. These answers are generally clear and relevant. \\
\hline
Medium &
  Answers at this level are somewhat accurate but may be incomplete or partially incorrect. They provide relevant information but may miss key details or present some minor inaccuracies. The response could be clearer or more comprehensive. \\
\hline
Bad &
  Answers in this category are largely inaccurate or irrelevant. They may partially address the question but contain significant errors or omissions. The response may be confusing, vague, or off-topic, requiring substantial correction or clarification. \\
\hline
Very Bad &
  Answers at this level are completely incorrect or irrelevant. They fail to address the question in any meaningful way, providing no useful information. The response might be entirely off-topic or nonsensical, reflecting a fundamental misunderstanding of the query. \\
\hline

\end{tabular}

\end{table}

\begin{table}[ht!]
\small\addtolength{\tabcolsep}{-5pt}
\hspace*{-0.3cm}
\caption{Percentage Data Quality}
\centering
\label{table:levels-of-data-quality}

\begin{tabular}{|p{0.7in}|p{0.7in}|p{0.7in}|}
\hline
  \textbf{Type} &
  \textbf{Number} & 
  \textbf{Percentage}\\
\hline
  Very good &
  631 &
  54.92 \%
  \\
\hline
Good &
  325 &
  28.28 \% \\
\hline
Medium &
  103 &
  8.96 \% \\
\hline
Bad &
  78  &
  6.78 \%\\
\hline
Very Bad &
  12 &
  1.05 \%\\
\hline
\textbf{Total} & 
    1149 & 
    100 \% \\  
\hline
\end{tabular}
\end{table}

\subsection{Result of model}
In this section, we present the performance of the Bloom and Vistral models. The results are evaluated using the training and validation loss metrics, as well as a comparison of the exact match (Exact) and F1 scores.

I implement hyperparameters with full fine-tuning model in table \ref{tab:hyper-peft} and hyperparameter using LoRA for tuning model in table \ref{tab:hyper-peft}.
\begin{table}[!hbt]
    \centering
    \caption{Hyperparameter of Bloom and Vistral models}
    \begin{tabular}{|c|c|c|}
        \hline
        \textbf{Model} &  \textbf{Bloom} & \textbf{Vistral} \\
        \hline
        $\beta_{1}$ & 0.9 & 0.9 \\
        \hline
        $\beta_{2}$ & 0.999 & 0.999 \\
        \hline
        \texttt{warmup ratio} & 0.05 & 0.05 \\
        \hline
        \texttt{weight decay} & 0.01 & 0.01\\
        \hline
        \texttt{batch size} & 8 & 4 \\
        \hline
        \texttt{max length} & 1024 & 1024 \\
        \hline
        \texttt{num epochs} & 10 & 10 \\
        \hline

    \end{tabular}
    
    \label{tab:hyper-non-peft}
\end{table}

\begin{table}[!hbt]
    \centering
    \caption{Hyperparameter of Bloom and Vistral models with LoRA}
    \begin{tabular}{|c|c|c|}
        \hline
        \textbf{Model} & \textbf{Bloom} & \textbf{Bloom} \\
        \hline
        $\beta_{1}$ & 0.9 & 0.9  \\
        \hline
        $\beta_{2}$ & 0.999 & 0.999 \\
        \hline
        \texttt{warmup ratio} & 0.05 & 0.05 \\
        \hline
        \texttt{weight decay} & 0.01 & 0.01 \\
        \hline
        \texttt{batch size} & 4 & 8 \\
        \hline
        \texttt{max length} & 1024 & 1024 \\
        \hline
        \texttt{num epochs} & 10 & 10 \\
        \hline
        \texttt{Rank LoRA} & 128 &  128 \\
        \hline
        \texttt{LoRA dropout} & 0.1 & 0.1 \\
        \hline
    \end{tabular}
    \label{tab:hyper-peft}
\end{table}

\begin{figure}[!hbt]
    \centering
    \includegraphics[width=0.95\linewidth]{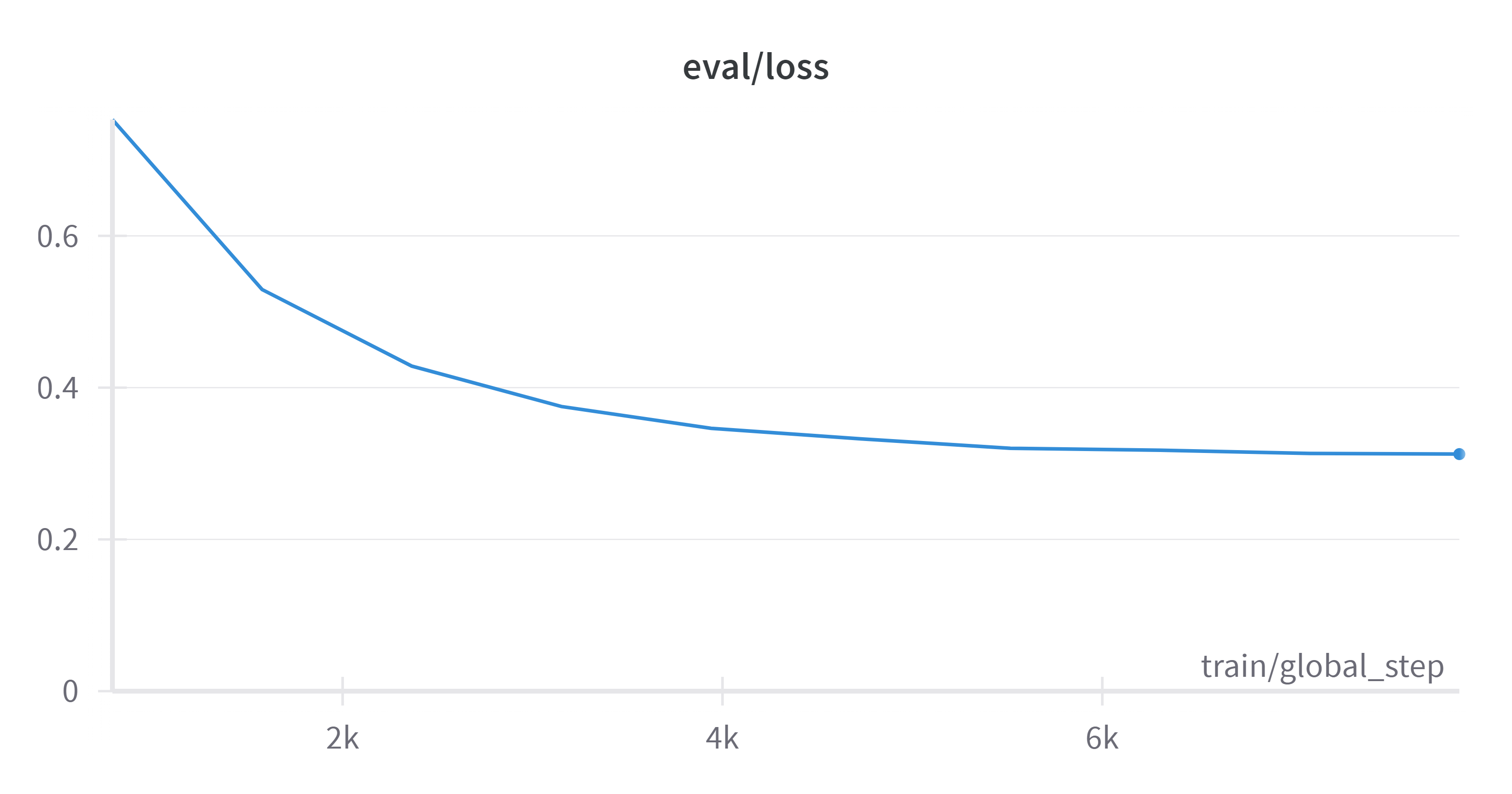}
    \caption{Validation Loss of Bloom Model}
    \label{fig:loss-eval-bloom}
\end{figure}

\begin{figure}[!hbt]
    \centering
    \includegraphics[width=0.95\linewidth]{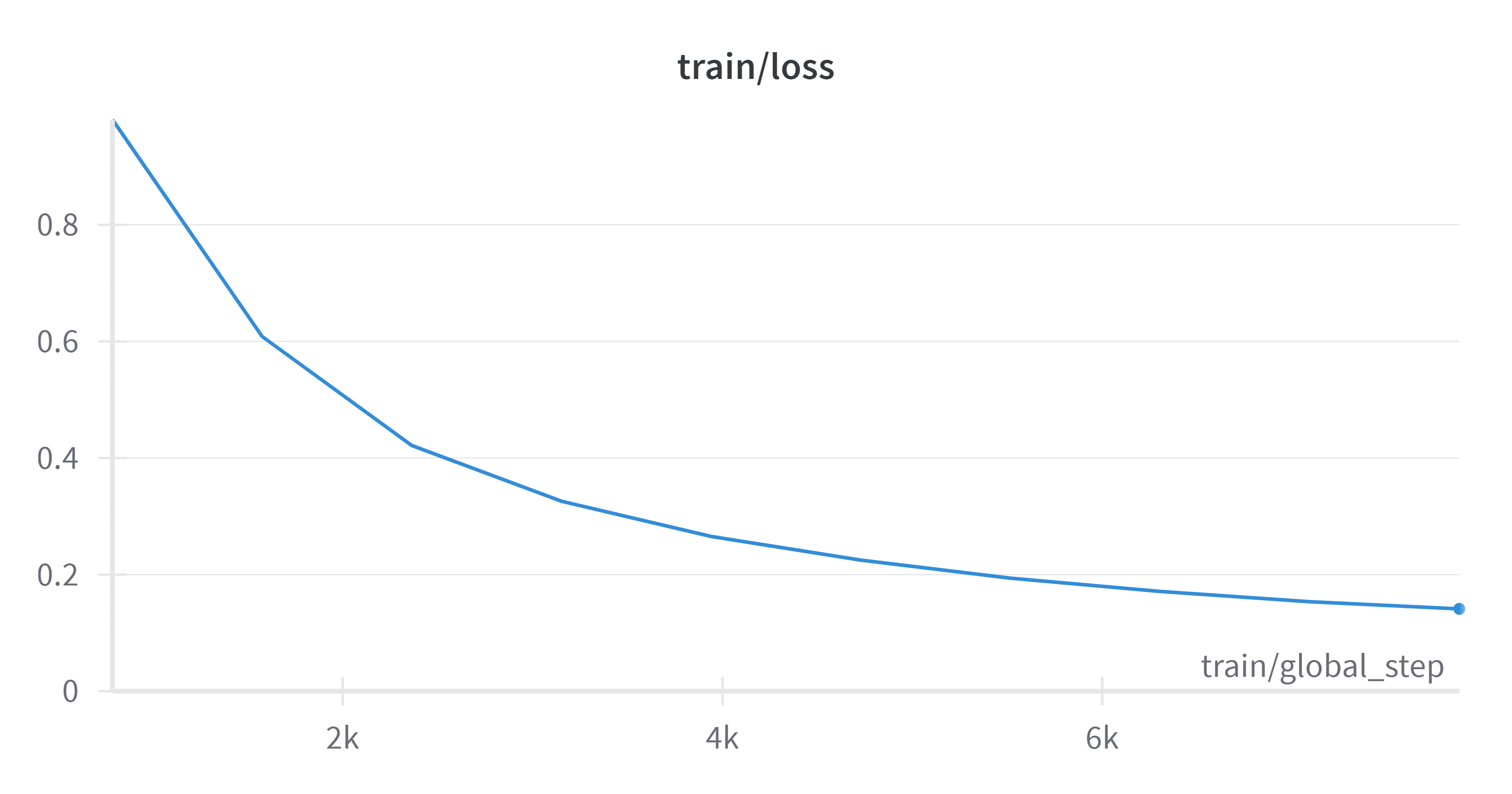}
    \caption{Training Loss of Bloom Model}
    \label{fig:loss-train-bloom-lora}
\end{figure}

\begin{figure}[!hbt]
    \centering
    \includegraphics[width=0.95\linewidth]{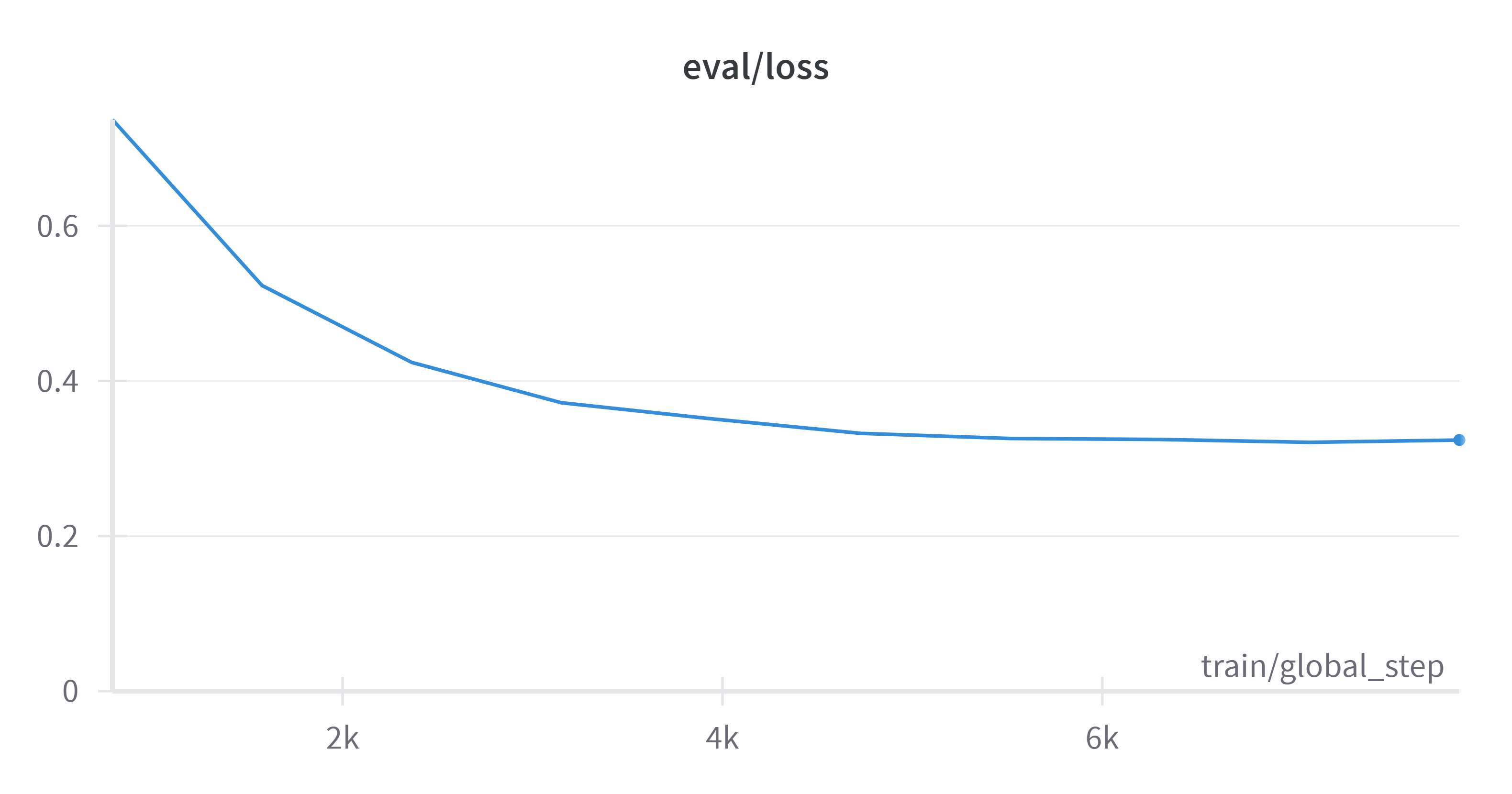}
    \caption{Validation Loss of Bloom Model + LoRA}
    \label{fig:loss-eval-bloom-lora}
\end{figure}

Training and Validation Loss
Bloom Model:

The training and validation loss curves for the Bloom model are shown in figures \ref{fig:lof-train-bloom} and \ref{fig:loss-eval-bloom}, respectively. Additionally, the training loss of Bloom model and LoRA method have training loss in figure \ref{fig:loss-train-bloom-lora} and validation loss illustrated in figure \ref{fig:loss-eval-bloom-lora}.

\begin{figure}[!hbt]
    \centering
    \includegraphics[width=0.95\linewidth]{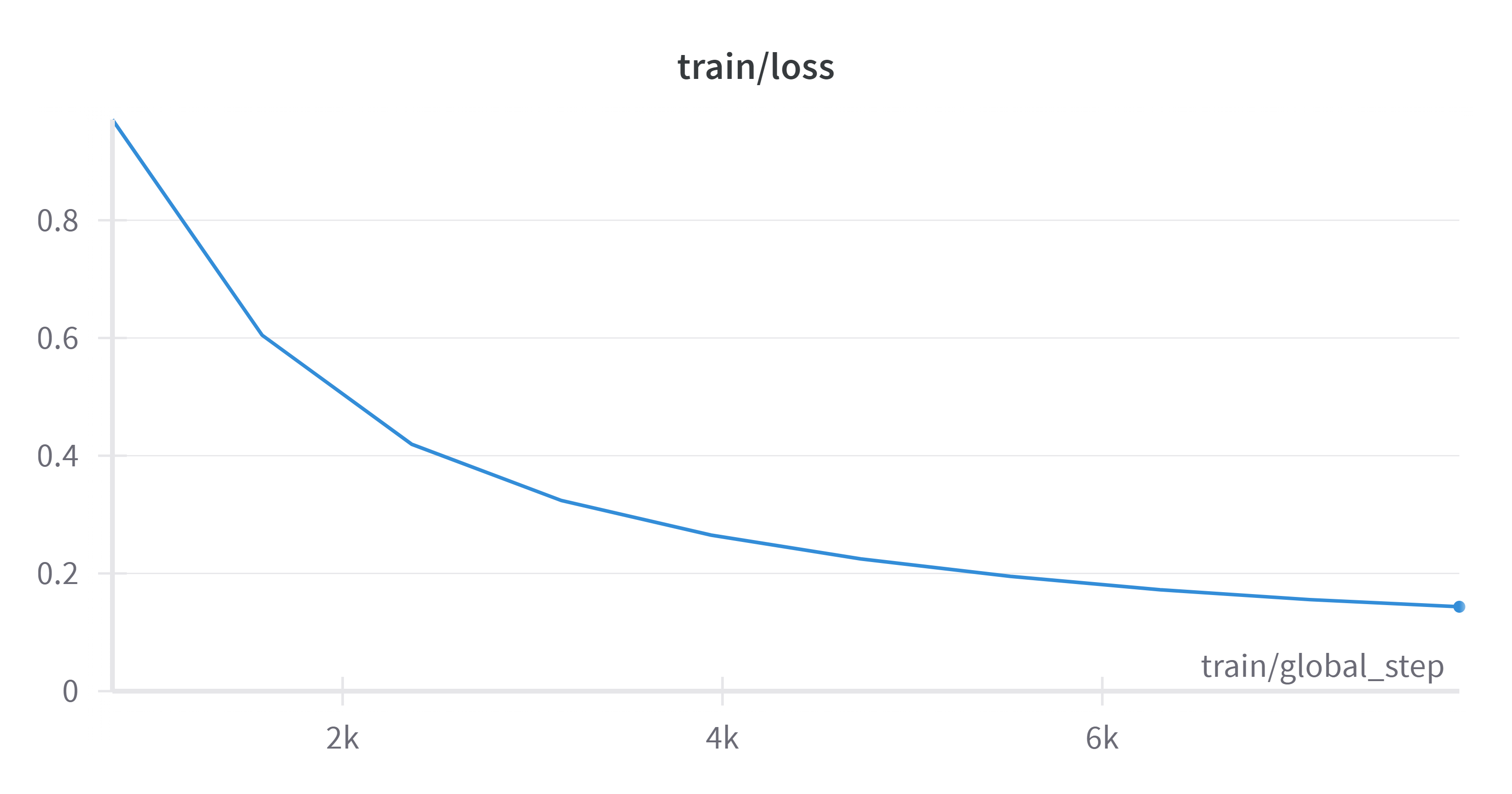}
    \caption{Training Loss of Bloom Model}
    \label{fig:loss-train-bloom}
\end{figure}

\begin{figure}[!hbt]
    \centering
    \includegraphics[width=0.95\linewidth]{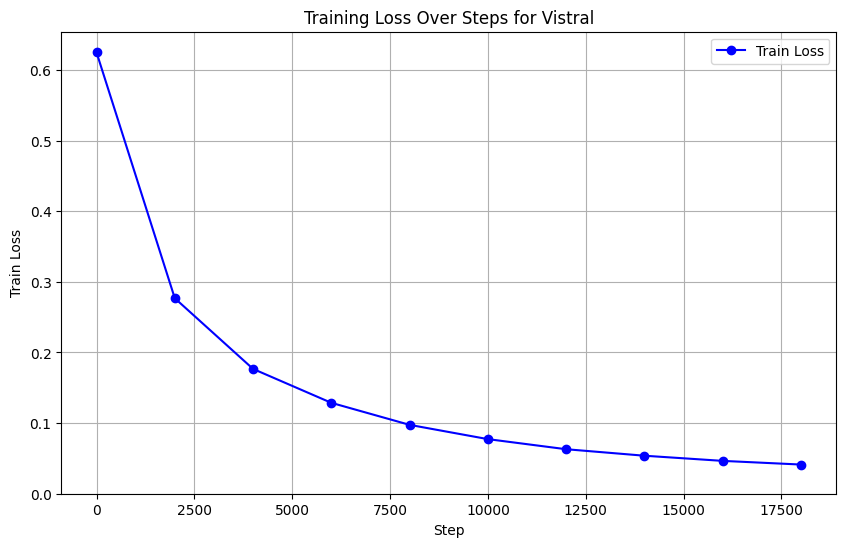}
    \caption{Training Loss of Vistral Model}
    \label{fig:loss-train-vistral}
\end{figure}

\begin{figure}[!hbt]
    \centering
    \includegraphics[width=0.95\linewidth]{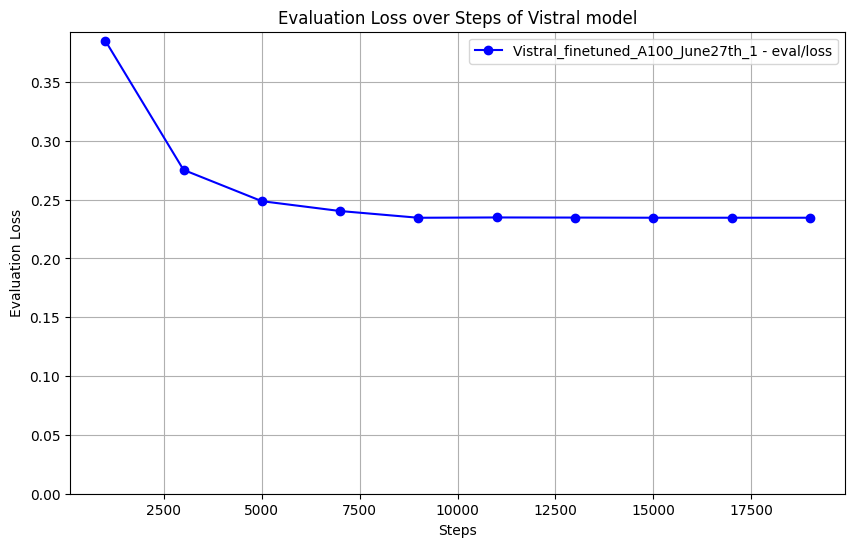}
    \caption{Validation Loss of Vistral Model}
    \label{fig:loss-eval-vistral}
\end{figure}

\begin{figure}[!hbt]
    \centering
    \includegraphics[width=0.95\linewidth]{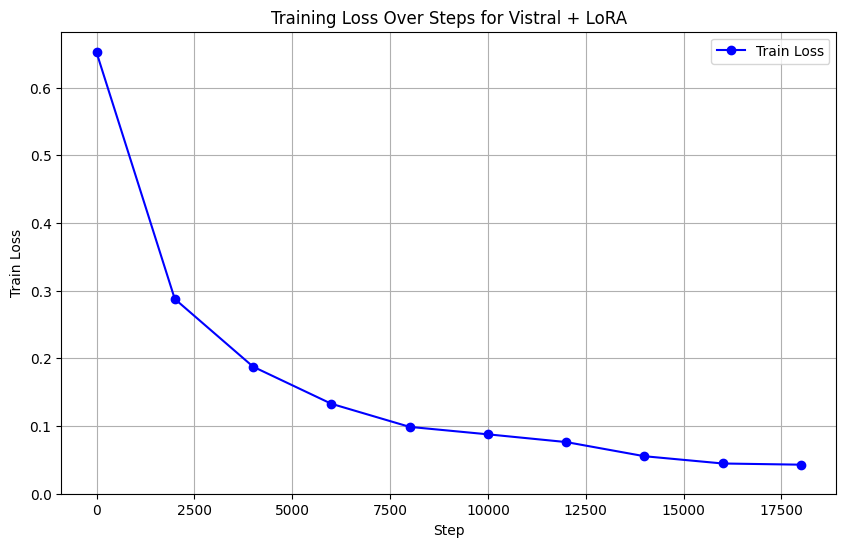}
    \caption{Training Loss of Vistral Model + LoRA}
    \label{fig:loss-train-vistral-lora}
\end{figure}

\begin{figure}[!hbt]
    \centering
    \includegraphics[width=0.95\linewidth]{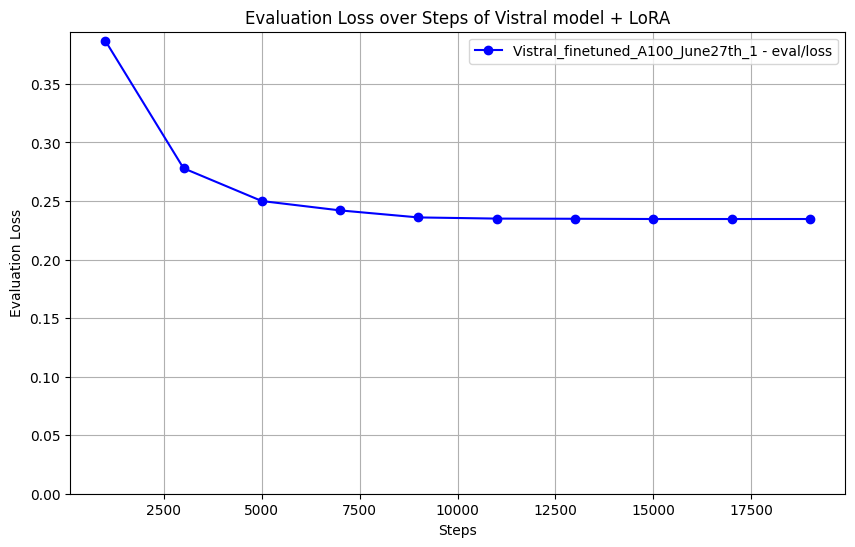}
    \caption{Validation Loss of Vistral Model + LoRA}
    \label{fig:loss-eval-vistral-lora}
\end{figure}

Vistral Model:

Similarly, the training and validation loss curves for the Vistral model are depicted in Figures \ref{fig:loss-train-vistral} and \ref{fig:loss-eval-vistral}. The Vistral model shows a rapid decrease in training loss, and the validation loss also reduces steadily, demonstrating good generalization performance.
Furthermore, in figure \ref{fig:loss-train-vistral-lora}, \ref{fig:loss-eval-vistral-lora} present loss of training and validate phrases respectively. 

Comparison of Bloom and Vistral Models
Table \ref{tab:result-bloom-vistral} provides a comparison of the Exact and F1 scores for both the Bloom and Vistral models. The Vistral model outperforms the Bloom model in both metrics, indicating its superior performance in terms of both accuracy and the quality of predictions.

\begin{table}[!hbt]
    \centering
    \caption{Overall result}
    \begin{tabular}{|c|c|c|}
        \hline
         \textbf{Metric} & \textbf{Exact} & \textbf{F1-score} \\
        \hline
         \textbf{Bloom model} + LoRA & 33.89 & 72.36 \\
         \hline
         \textbf{Vistral} + LoRA & 43.23 & 81.24 \\
         \hline
         \textbf{Bloom model} & 34.23 & 73.16 \\
         \hline
         \textbf{Vistral model} & 43.72 & 81.57 \\
         \hline
    \end{tabular}
    
    \label{tab:result-bloom-vistral}
\end{table}

\begin{table}[!hbt]
    \centering
    \caption{Resource usage of language models}
    \begin{tabular}{|p{0.5in}|p{0.8in}|p{0.7in}|}
        \hline
         \textbf{Model} & \textbf{Time training per epoch} & \textbf{Ram-GPU used} \\
        \hline
         \textbf{Bloom model} + LoRA & 1.5 hours & 16 GB \\
         \hline
         \textbf{Vistral} + LoRA & 6 hours & 32 GB \\
         \hline
         \textbf{Bloom model} & 5 hours & 29 GB \\
         \hline
         \textbf{Vistral model} & 14 hours & 61.2 GB \\
         \hline
    \end{tabular}
    \label{tab:result-bloom-vistral-lora}
\end{table}

%%%

\section{Analysis and discussion}
\subsection{Performance Metrics}
\begin{itemize}
    \item Bloom model + LoRA vs. Bloom model: The Bloom model with LoRA shows a slight decrease in Exact and F1-score compared to the Bloom model without LoRA. The Exact score drops from 34.23 to 33.89, and the F1 score decreases from 73.16 to 72.36. This suggests that LoRA might slightly affect the performance of the Bloom model in terms of these metrics.

    \item Vistral + LoRA vs. Vistral: The Vistral model with LoRA also exhibits a minor reduction in performance compared to the Vistral model without LoRA. The Exact score drops from 43.72 to 43.23, and the F1 score decreases from 81.57 to 81.24. This indicates that the inclusion of LoRA may have a small impact on the Vistral model's performance.

    \item Bloom model vs. Vistral: Comparing the two models, Vistral consistently outperforms Bloom in both Exact and F1-score, both with and without LoRA. This demonstrates that the Vistral model is more effective in capturing and processing the information needed for higher precision and overall accuracy.
\end{itemize}

\subsection{Resource Utilization}
\begin{itemize}
    \item Training Time: The training time per epoch is significantly lower for models using LoRA. The Bloom model with LoRA takes 1.5 hours per epoch, whereas without LoRA, it takes 5 hours. Similarly, the Vistral model with LoRA takes 6 hours per epoch, compared to 14 hours without LoRA. This reduction in training time highlights the efficiency of the LoRA method in speeding up the training process.

    \item GPU RAM Usage: Models with LoRA also require less GPU RAM. The Bloom model with LoRA uses 16 GB, while the original Bloom model uses 29 GB. The Vistral model with LoRA uses 32 GB, compared to 61.2 GB for the Vistral model without LoRA. This reduction in memory usage indicates that LoRA helps in optimizing resource utilization during training.
\end{itemize}

\subsection{Real-world inference}
Example in table \ref{tab:good-answer} there are four reasons explain why this is a good answer:
\begin{itemize}
    \item Comprehensive and Detailed: The answer includes the main responsibilities of the lecturers such as imparting scientific ambition, a passion for learning, scientific research, scientific thinking, and creative ability. These elements are clearly stated in the context provided.

    \item Clear and Understandable: The answer is articulated clearly and understandably, making it easy for the reader to grasp the responsibilities of the lecturers.

    \item Contextual Connection: The answer is closely linked to the context provided, ensuring that the information presented is accurate and relevant.

    \item Repetition for Emphasis: Repeating the key points at the end of the answer emphasizes the lecturers' responsibilities and highlights the main aspects that need to be noted.
\end{itemize}

\begin{table}[!hbt]
    \centering
    \caption{Good answer in model}
    \begin{tabular}{|p{0.5 in}|p{2.2in}|}
    \hline
         \textbf{Context} & Điều 33. GIảng viên
c ) Ngoài những nghĩa vụ chung đối với giảng viên , các giảng viên tham gia đào tạo tài năng , chất lượng cao còn có trách nhiệm sau :
- Truyền thụ và tạo lập cho sinh viên hoài bão khoa học , niềm say mê học tập , nghiên cứu khoa học ,tư duy khoa học và năng lực sáng tạo ,động viên ,khuyến khích sinh viên rèn luyện phẩm chất chính trị,đạo đức tác phong của người cán bộ nghiên cứu khoa học ;
- Tự giác áp dụng và hướng dẫn sinh viên áp dụng phương pháp dạy -học , kiểm tra đánh giá tiên tiến phù hợp ;
- Hướng dẫn sinh viên nghiên cứu khoa học ;
Cung cấp học liệu cho sinh viên trước giờ lên lớp ở một trong các dạng sau : giáo trình ;bài giảng ; đề cương chi tiết bài giảng cùng danh mục tài liệu tham khảo có trong thư viện . Nội dung bài giảng phải cơ bản , hiện đại, trình bày được các thành tựu nghiên cứu mới trong nước và quốc tế,phù hợp với mục tiêu đào tạo , giới thiệu các tàiliệu tham khảo phong phú cho sinh viên.
"
 \\
 \hline
       \textbf{Question}  & Giảng viên có trách nhiệm gì đối với sinh viên trong quá trình đào tạo? \\
       \hline
       \textbf{Answer} & Giảng viên có trách nhiệm truyền thụ và tạo lập cho sinh viên hoài bão khoa học, niềm say mê học tập, nghiên cứu khoa học, tư duy khoa học và năng lực sáng tạo. Hãy suy nghĩ từng bước. Vì vậy, câu trả lời là giảng viên có trách nhiệm truyền thụ và tạo lập cho sinh viên hoài bão khoa học, niềm say mê học tập, nghiên cứu khoa học, tư duy khoa học và năng lực sáng tạo.
        \\
        \hline
    \end{tabular}
    
    \label{tab:good-answer}
\end{table}

In table \ref{tab:bad-answer} illustrates the bad answer. This poor answer does not provide a comprehensive response based on the provided context. It lacks details and does not address the specific aspects mentioned in the context, such as training programs, study time, organization and management of training, scientific research, rights and obligations of lecturers, advisors, students, examinations, assessments, and graduation recognition.
\begin{table}[!hbt]
    \centering
    \caption{Bad answer in model}
    \begin{tabular}{|p{0.5 in}|p{2.2in}|}
    \hline
         \textbf{Context} & Điều 1. Phạm vi điều chỉnh và đối tượng áp dụng
1. Quy chế này quy định về đào tạo đại học theo hình thức chính quy và hình thức vừa làm vừa học tại Đại học Quốc gia Hà Nội ,bao gồm :chương trình đào tạo và thời gian học tập ;tổ chức và quản lý đào tạo; nghiên cứu khoa học của sinh viên ; quyền lợi và nghĩa vụ của giảng viên , cố vấn học tập, sinh viên ; kiểm tra, thi và đánh giá kết quả học tập ;công nhận tốt nghiệp

 \\
 \hline
       \textbf{Question}  & Quy chế đào tạo đại học quy định về những gì?
 \\
       \hline
       \textbf{Answer} & Quy chế đào tạo đại học quy định về đào tạo.
        \\
        \hline
    \end{tabular}
    
    \label{tab:bad-answer}
\end{table}

%%%
\section*{Conclusion and Limitations}
% \section{Conclusion and Limitation}
\subsection{Conclusion}

In this paper, we present a simple and effective framework for applying large language models (LLMs) to educational domain. We conduct the experiments with fine-tuning methods on resource-constrained environments to optimally leverage existing GPU capabilities and hardware. Our results demonstrated that using LLMs models for vietnamese improved performance by over 10 points compared to previous model. This significant improvement highlights the effectiveness of our approach in maximizing the potential of limited computational resources.

\subsection{Limitations}
In this study and in the realm of natural language processing, particularly in the application of question-answering (QA) systems for educational management in Vietnamese, several limitations of current models and data quality have been identified. These limitations are crucial to understand for the continued development and improvement of such systems.

\begin{enumerate}
    \item \textbf{Reasoning Capabilities of the Model} 
    \begin{itemize}
        \item \textbf{Logical Reasoning:} The models may produce answers that lack coherent logical structure or fail to follow a clear line of reasoning, especially for complex or multi-step problems.
        \item \textbf{Contextual Understanding:} While models can understand the context to a certain extent, they often miss subtle nuances and deeper connections within the provided context, leading to less accurate or irrelevant responses.
    \end{itemize}

    \item \textbf{Contextual Errors and Ambiguity}
    \begin{itemize}
        \item \textbf{Error in Capturing Context:} Models sometimes fail to capture the full context of a question, particularly when the context is lengthy or contains intricate details.
        \item \textbf{Ambiguity in Responses:} Due to the models' probabilistic nature, they can produce responses that are ambiguous or vague, which can be particularly problematic in educational management where precision is crucial.
    \end{itemize}

    \item \textbf{Lack of Specialized Knowledge}
    \begin{itemize}
        \item \textbf{Handling Specific Regulations:} The models might not fully grasp the specific regulations and guidelines unique to different educational institutions or contexts, leading to incorrect or incomplete answers.
        \item \textbf{Domain-Specific Expertise:} The absence of deep domain expertise means that the models might misinterpret or overlook critical aspects of educational management tasks.
    \end{itemize}
 
\end{enumerate}

\bibliography{custom}

\begin{thebibliography}{62}
\providecommand{\natexlab}[1]{#1}

\bibitem[{Agrawal et~al.(2015)Agrawal, Lu, Antol, Mitchell, Zitnick, Parikh, and Batra}]{Agrawal2015VQAVQ}
Aishwarya Agrawal, Jiasen Lu, Stanislaw Antol, Margaret Mitchell, C.~Lawrence Zitnick, Devi Parikh, and Dhruv Batra. 2015.
\newblock \href {https://api.semanticscholar.org/CorpusID:3180429} {Vqa: Visual question answering}.
\newblock \emph{International Journal of Computer Vision}, 123:4 -- 31.

\bibitem[{Andreas et~al.(2016)Andreas, Rohrbach, Darrell, and Klein}]{Andreas2016LearningTC}
Jacob Andreas, Marcus Rohrbach, Trevor Darrell, and Dan Klein. 2016.
\newblock \href {https://api.semanticscholar.org/CorpusID:3130692} {Learning to compose neural networks for question answering}.
\newblock \emph{ArXiv}, abs/1601.01705.

\bibitem[{avelka et~al.(2023)avelka, Agarwal, An, Bogart, and Sakr}]{avelka2023ThrilledBY}
Jaromr avelka, Arav Agarwal, Marshall An, Christopher Bogart, and Majd~F. Sakr. 2023.
\newblock \href {https://api.semanticscholar.org/CorpusID:259203041} {Thrilled by your progress! large language models (gpt-4) no longer struggle to pass assessments in higher education programming courses}.
\newblock \emph{Proceedings of the 2023 ACM Conference on International Computing Education Research - Volume 1}.

\bibitem[{Batool et~al.(2022)Batool, Rashid, Nisar, Kim, Kwon, and Hussain}]{Batool2022EducationalDM}
Saba Batool, Junaid Rashid, Muhammad~Wasif Nisar, Jungeun Kim, Hyuk-Yoon Kwon, and Amir Hussain. 2022.
\newblock \href {https://api.semanticscholar.org/CorpusID:250417371} {Educational data mining to predict students' academic performance: A survey study}.
\newblock \emph{Education and Information Technologies}, 28:905--971.

\bibitem[{Bhowmick et~al.(2023)Bhowmick, Jagmohan, Vempaty, Dey, Hall, Hartman, Kokku, and Maheshwari}]{Bhowmick2023AutomatingQG}
Ayan~Kumar Bhowmick, Ashish Jagmohan, Aditya Vempaty, Prasenjit Dey, Leigh Ann~Edwards Hall, Jeremy Hartman, Ravi Kokku, and Hema Maheshwari. 2023.
\newblock \href {https://api.semanticscholar.org/CorpusID:262824938} {Automating question generation from educational text}.
\newblock In \emph{SGAI Conferences}.

\bibitem[{Bommarito and Katz(2022)}]{Bommarito2022GPTTT}
Michael~James Bommarito and Daniel~Martin Katz. 2022.
\newblock \href {https://api.semanticscholar.org/CorpusID:255340451} {Gpt takes the bar exam}.
\newblock \emph{ArXiv}, abs/2212.14402.

\bibitem[{Chen et~al.(2020)Chen, Chen, and Lin}]{Chen2020ArtificialII}
Lijia Chen, Pingping Chen, and Zhijian Lin. 2020.
\newblock \href {https://api.semanticscholar.org/CorpusID:218493891} {Artificial intelligence in education: A review}.
\newblock \emph{IEEE Access}, 8:75264--75278.

\bibitem[{Chen et~al.(2022)Chen, Ma, Wang, and Cohen}]{Chen2022ProgramOT}
Wenhu Chen, Xueguang Ma, Xinyi Wang, and William~W. Cohen. 2022.
\newblock \href {https://api.semanticscholar.org/CorpusID:253801709} {Program of thoughts prompting: Disentangling computation from reasoning for numerical reasoning tasks}.
\newblock \emph{ArXiv}, abs/2211.12588.

\bibitem[{Chen et~al.(2023{\natexlab{a}})Chen, Yin, Ku, Wan, Ma, Xu, Xia, Wang, and Lu}]{Chen2023TheoremQAAT}
Wenhu Chen, Ming Yin, Max~W.F. Ku, Yixin Wan, Xueguang Ma, Jianyu Xu, Tony Xia, Xinyi Wang, and Pan Lu. 2023{\natexlab{a}}.
\newblock \href {https://api.semanticscholar.org/CorpusID:258833200} {Theoremqa: A theorem-driven question answering dataset}.
\newblock \emph{ArXiv}, abs/2305.12524.

\bibitem[{Chen et~al.(2023{\natexlab{b}})Chen, Mao, Li, Jin, Wen, Wei, Wang, Yin, Fan, Liu, and Tang}]{Chen2023ExploringTP}
Zhikai Chen, Haitao Mao, Hang Li, Wei Jin, Haifang Wen, Xiaochi Wei, Shuaiqiang Wang, Dawei Yin, Wenqi Fan, Hui Liu, and Jiliang Tang. 2023{\natexlab{b}}.
\newblock \href {https://api.semanticscholar.org/CorpusID:259375824} {Exploring the potential of large language models (llms)in learning on graphs}.
\newblock \emph{ACM SIGKDD Explorations Newsletter}, 25:42 -- 61.

\bibitem[{Cui et~al.(2023)Cui, Li, Yan, Chen, and Yuan}]{Cui2023ChatlawAM}
Jiaxi Cui, Zongjia Li, Yang Yan, Bohua Chen, and Li~Yuan. 2023.
\newblock \href {https://api.semanticscholar.org/CorpusID:259274889} {Chatlaw: A multi-agent collaborative legal assistant with knowledge graph enhanced mixture-of-experts large language model}.

\bibitem[{Denny et~al.(2023)Denny, Prather, Becker, Finnie-Ansley, Hellas, Leinonen, Luxton-Reilly, Reeves, Santos, and Sarsa}]{Denny2023ComputingEI}
Paul Denny, James Prather, Brett~A. Becker, James Finnie-Ansley, Arto Hellas, Juho Leinonen, Andrew Luxton-Reilly, Brent~N. Reeves, Eddie~Antonio Santos, and Sami Sarsa. 2023.
\newblock \href {https://api.semanticscholar.org/CorpusID:259076191} {Computing education in the era of generative ai}.
\newblock \emph{Communications of the ACM}, 67:56 -- 67.

\bibitem[{Dewhurst et~al.(2000)Dewhurst, Macleod, and Norris}]{Dewhurst2000IndependentSL}
David Dewhurst, Hamish Macleod, and Tracey A.~M. Norris. 2000.
\newblock \href {https://api.semanticscholar.org/CorpusID:206063996} {Independent student learning aided by computers: an acceptable alternative to lectures?}
\newblock \emph{Comput. Educ.}, 35:223--241.

\bibitem[{Dosovitskiy et~al.(2020)Dosovitskiy, Beyer, Kolesnikov, Weissenborn, Zhai, Unterthiner, Dehghani, Minderer, Heigold, Gelly, Uszkoreit, and Houlsby}]{Dosovitskiy2020AnII}
Alexey Dosovitskiy, Lucas Beyer, Alexander Kolesnikov, Dirk Weissenborn, Xiaohua Zhai, Thomas Unterthiner, Mostafa Dehghani, Matthias Minderer, Georg Heigold, Sylvain Gelly, Jakob Uszkoreit, and Neil Houlsby. 2020.
\newblock \href {https://api.semanticscholar.org/CorpusID:225039882} {An image is worth 16x16 words: Transformers for image recognition at scale}.
\newblock \emph{ArXiv}, abs/2010.11929.

\bibitem[{Fan et~al.(2023)Fan, Zhao, Li, Liu, Mei, Wang, Tang, and Li}]{Fan2023RecommenderSI}
Wenqi Fan, Zihuai Zhao, Jiatong Li, Yunqing Liu, Xiaowei Mei, Yiqi Wang, Jiliang Tang, and Qing Li. 2023.
\newblock \href {https://api.semanticscholar.org/CorpusID:259342486} {Recommender systems in the era of large language models (llms)}.
\newblock \emph{ArXiv}, abs/2307.02046.

\bibitem[{Fukui et~al.(2016)Fukui, Park, Yang, Rohrbach, Darrell, and Rohrbach}]{Fukui2016MultimodalCB}
Akira Fukui, Dong~Huk Park, Daylen Yang, Anna Rohrbach, Trevor Darrell, and Marcus Rohrbach. 2016.
\newblock \href {https://api.semanticscholar.org/CorpusID:2840197} {Multimodal compact bilinear pooling for visual question answering and visual grounding}.
\newblock In \emph{Conference on Empirical Methods in Natural Language Processing}.

\bibitem[{Gao et~al.(2022{\natexlab{a}})Gao, Ping, Thattai, Reganti, Wu, and Natarajan}]{Gao2022TransformRetrieveGenerateNL}
Feng Gao, Q.~Ping, Govind Thattai, Aishwarya~N. Reganti, Yingting Wu, and Premkumar Natarajan. 2022{\natexlab{a}}.
\newblock \href {https://api.semanticscholar.org/CorpusID:249455925} {Transform-retrieve-generate: Natural language-centric outside-knowledge visual question answering}.
\newblock \emph{2022 IEEE/CVF Conference on Computer Vision and Pattern Recognition (CVPR)}, pages 5057--5067.

\bibitem[{Gao et~al.(2022{\natexlab{b}})Gao, Madaan, Zhou, Alon, Liu, Yang, Callan, and Neubig}]{Gao2022PALPL}
Luyu Gao, Aman Madaan, Shuyan Zhou, Uri Alon, Pengfei Liu, Yiming Yang, Jamie Callan, and Graham Neubig. 2022{\natexlab{b}}.
\newblock \href {https://api.semanticscholar.org/CorpusID:253708270} {Pal: Program-aided language models}.
\newblock \emph{ArXiv}, abs/2211.10435.

\bibitem[{Gao et~al.(2018)Gao, Jiang, You, Lu, Hoi, Wang, and Li}]{Gao2018DynamicFW}
Peng Gao, Zhengkai Jiang, Haoxuan You, Pan Lu, Steven C.~H. Hoi, Xiaogang Wang, and Hongsheng Li. 2018.
\newblock \href {https://api.semanticscholar.org/CorpusID:54700454} {Dynamic fusion with intra- and inter-modality attention flow for visual question answering}.
\newblock \emph{2019 IEEE/CVF Conference on Computer Vision and Pattern Recognition (CVPR)}, pages 6632--6641.

\bibitem[{Jin et~al.(2020)Jin, Pan, Oufattole, Weng, Fang, and Szolovits}]{Jin2020WhatDD}
Di~Jin, Eileen Pan, Nassim Oufattole, Wei-Hung Weng, Hanyi Fang, and Peter Szolovits. 2020.
\newblock \href {https://api.semanticscholar.org/CorpusID:221970190} {What disease does this patient have? a large-scale open domain question answering dataset from medical exams}.
\newblock \emph{ArXiv}, abs/2009.13081.

\bibitem[{Jin et~al.(2024)Jin, Zhang, Chen, Zhang, Liang, Yang, Wang, Pan, and Wen}]{Jin2024PositionWC}
Ming Jin, Yifan Zhang, Wei Chen, Kexin Zhang, Yuxuan Liang, Bin Yang, Jindong Wang, Shirui Pan, and Qingsong Wen. 2024.
\newblock \href {https://api.semanticscholar.org/CorpusID:270213739} {Position: What can large language models tell us about time series analysis}.

\bibitem[{Kafle et~al.(2018)Kafle, Cohen, Price, and Kanan}]{Kafle2018DVQAUD}
Kushal Kafle, Scott~D. Cohen, Brian~L. Price, and Christopher Kanan. 2018.
\newblock \href {https://api.semanticscholar.org/CorpusID:4445015} {Dvqa: Understanding data visualizations via question answering}.
\newblock \emph{2018 IEEE/CVF Conference on Computer Vision and Pattern Recognition}, pages 5648--5656.

\bibitem[{Kasneci et~al.(2023)Kasneci, Se{\ss}ler, K{\"u}chemann, Bannert, Dementieva, Fischer, Gasser, Groh, G{\"u}nnemann, H{\"u}llermeier, Krusche, Kutyniok, Michaeli, Nerdel, Pfeffer, Poquet, Sailer, Schmidt, Seidel, Stadler, Weller, Kuhn, and Kasneci}]{Kasneci2023ChatGPTFG}
Enkelejda Kasneci, Kathrin Se{\ss}ler, Stefan K{\"u}chemann, Maria Bannert, Daryna Dementieva, Frank Fischer, Urs Gasser, George~Louis Groh, Stephan G{\"u}nnemann, Eyke H{\"u}llermeier, Stephan Krusche, Gitta Kutyniok, Tilman Michaeli, Claudia Nerdel, J{\"u}rgen Pfeffer, Oleksandra Poquet, Michael Sailer, Albrecht Schmidt, Tina Seidel, Matthias Stadler, Jochen Weller, Jochen Kuhn, and Gjergji Kasneci. 2023.
\newblock \href {https://api.semanticscholar.org/CorpusID:257445349} {Chatgpt for good? on opportunities and challenges of large language models for education}.
\newblock \emph{Learning and Individual Differences}.

\bibitem[{Kazemitabaar et~al.(2023)Kazemitabaar, Hou, Henley, Ericson, Weintrop, and Grossman}]{Kazemitabaar2023HowNU}
Majeed Kazemitabaar, Xinying Hou, Austin Henley, Barb Ericson, David Weintrop, and Tovi Grossman. 2023.
\newblock \href {https://api.semanticscholar.org/CorpusID:262459210} {How novices use llm-based code generators to solve cs1 coding tasks in a self-paced learning environment}.
\newblock \emph{Proceedings of the 23rd Koli Calling International Conference on Computing Education Research}.

\bibitem[{Kembhavi et~al.(2016)Kembhavi, Salvato, Kolve, Seo, Hajishirzi, and Farhadi}]{Kembhavi2016ADI}
Aniruddha Kembhavi, Michael Salvato, Eric Kolve, Minjoon Seo, Hannaneh Hajishirzi, and Ali Farhadi. 2016.
\newblock \href {https://api.semanticscholar.org/CorpusID:2682274} {A diagram is worth a dozen images}.
\newblock \emph{ArXiv}, abs/1603.07396.

\bibitem[{Kembhavi et~al.(2017)Kembhavi, Seo, Schwenk, Choi, Farhadi, and Hajishirzi}]{Kembhavi2017AreYS}
Aniruddha Kembhavi, Minjoon Seo, Dustin Schwenk, Jonghyun Choi, Ali Farhadi, and Hannaneh Hajishirzi. 2017.
\newblock \href {https://api.semanticscholar.org/CorpusID:1310550} {Are you smarter than a sixth grader? textbook question answering for multimodal machine comprehension}.
\newblock \emph{2017 IEEE Conference on Computer Vision and Pattern Recognition (CVPR)}, pages 5376--5384.

\bibitem[{Koedinger et~al.(2015)Koedinger, D’Mello, Mclaughlin, Pardos, and Ros{\'e}}]{Koedinger2015DataMA}
K.~Koedinger, Sidney~K. D’Mello, Elizabeth Mclaughlin, Zachary~A. Pardos, and Carolyn~Penstein Ros{\'e}. 2015.
\newblock \href {https://api.semanticscholar.org/CorpusID:26099556} {Data mining and education.}
\newblock \emph{Wiley interdisciplinary reviews. Cognitive science}, 6 4:333--353.

\bibitem[{K{\"o}nig et~al.(2022)K{\"o}nig, Karrenbauer, and Breitner}]{Knig2022CriticalSF}
Claudia~M. K{\"o}nig, Christin Karrenbauer, and Michael~H. Breitner. 2022.
\newblock \href {https://api.semanticscholar.org/CorpusID:253019133} {Critical success factors and challenges for individual digital study assistants in higher education: A mixed methods analysis}.
\newblock \emph{Education and Information Technologies}, 28:4475 -- 4503.

\bibitem[{Latif et~al.(2023)Latif, Mai, Nyaaba, Wu, Liu, Lu, Li, Liu, and Zhai}]{Latif2023ArtificialGI}
Ehsan Latif, Gengchen Mai, Matthew Nyaaba, Xuansheng Wu, Ninghao Liu, Guoyu Lu, Sheng Li, Tianming Liu, and Xiaoming Zhai. 2023.
\newblock \href {https://api.semanticscholar.org/CorpusID:267749183} {Artificial general intelligence (agi) for education}.
\newblock \emph{ArXiv}, abs/2304.12479.

\bibitem[{Lee et~al.(2019)Lee, Kim, and Park}]{Lee2019DomainagnosticQW}
Seanie Lee, Donggyu Kim, and Jangwon Park. 2019.
\newblock \href {https://api.semanticscholar.org/CorpusID:204800552} {Domain-agnostic question-answering with adversarial training}.
\newblock In \emph{Conference on Empirical Methods in Natural Language Processing}.

\bibitem[{Lende and Raghuwanshi(2016)}]{Lende2016QuestionAS}
Sweta~P. Lende and Mukesh~M. Raghuwanshi. 2016.
\newblock \href {https://api.semanticscholar.org/CorpusID:32198394} {Question answering system on education acts using nlp techniques}.
\newblock \emph{2016 World Conference on Futuristic Trends in Research and Innovation for Social Welfare (Startup Conclave)}, pages 1--6.

\bibitem[{Li et~al.(2024)Li, Xu, Zhang, Chen, Liang, Fan, Li, Tang, and Wen}]{Li2024BringingGA}
Hang Li, Tianlong Xu, Chaoli Zhang, Eason Chen, Jing Liang, Xing Fan, Haoyang Li, Jiliang Tang, and Qingsong Wen. 2024.
\newblock \href {https://api.semanticscholar.org/CorpusID:267783076} {Bringing generative ai to adaptive learning in education}.
\newblock \emph{ArXiv}, abs/2402.14601.

\bibitem[{Li et~al.(2023)Li, Fu, Zhang, Chen, Yu, Xia, Zhang, Tang, and Yu}]{Li2023AdaptingLL}
Qingyao Li, Lingyue Fu, Weiming Zhang, Xianyu Chen, Jingwei Yu, Wei Xia, Weinan Zhang, Ruiming Tang, and Yong Yu. 2023.
\newblock \href {https://api.semanticscholar.org/CorpusID:267027689} {Adapting large language models for education: Foundational capabilities, potentials, and challenges}.
\newblock \emph{ArXiv}, abs/2401.08664.

\bibitem[{Li'evin et~al.(2022)Li'evin, Hother, and Winther}]{Lievin2022CanLL}
Valentin Li'evin, Christoffer~Egeberg Hother, and Ole Winther. 2022.
\newblock \href {https://api.semanticscholar.org/CorpusID:250627547} {Can large language models reason about medical questions?}
\newblock \emph{Patterns}, 5.

\bibitem[{Liu et~al.(2022)Liu, Chen, Li, Tan, Cai, and Ayub}]{Liu2022APM}
Fen Liu, Jianfeng Chen, Kemeng Li, Weijie Tan, Chang Cai, and Muhammad~Saad Ayub. 2022.
\newblock \href {https://api.semanticscholar.org/CorpusID:254900038} {A parallel multi-modal factorized bilinear pooling fusion method based on the semi-tensor product for emotion recognition}.
\newblock \emph{Entropy}, 24.

\bibitem[{Lu et~al.(2016)Lu, Yang, Batra, and Parikh}]{Lu2016HierarchicalQC}
Jiasen Lu, Jianwei Yang, Dhruv Batra, and Devi Parikh. 2016.
\newblock \href {https://api.semanticscholar.org/CorpusID:868693} {Hierarchical question-image co-attention for visual question answering}.
\newblock \emph{ArXiv}, abs/1606.00061.

\bibitem[{Lu et~al.(2022)Lu, Mishra, Xia, Qiu, Chang, Zhu, Tafjord, Clark, and Kalyan}]{Lu2022LearnTE}
Pan Lu, Swaroop Mishra, Tony Xia, Liang Qiu, Kai-Wei Chang, Song-Chun Zhu, Oyvind Tafjord, Peter Clark, and A.~Kalyan. 2022.
\newblock \href {https://api.semanticscholar.org/CorpusID:252383606} {Learn to explain: Multimodal reasoning via thought chains for science question answering}.
\newblock \emph{ArXiv}, abs/2209.09513.

\bibitem[{Malinka et~al.(2023)Malinka, Peres{\'i}ni, Firc, Hujň{\'a}k, and Janus}]{Malinka2023OnTE}
Kamil Malinka, Martin Peres{\'i}ni, Anton Firc, Ondřej Hujň{\'a}k, and Filip Janus. 2023.
\newblock \href {https://api.semanticscholar.org/CorpusID:257631490} {On the educational impact of chatgpt: Is artificial intelligence ready to obtain a university degree?}
\newblock \emph{Proceedings of the 2023 Conference on Innovation and Technology in Computer Science Education V. 1}.

\bibitem[{Malinowski et~al.(2015)Malinowski, Rohrbach, and Fritz}]{Malinowski2015AskYN}
Mateusz Malinowski, Marcus Rohrbach, and Mario Fritz. 2015.
\newblock \href {https://api.semanticscholar.org/CorpusID:738850} {Ask your neurons: A neural-based approach to answering questions about images}.
\newblock \emph{2015 IEEE International Conference on Computer Vision (ICCV)}, pages 1--9.

\bibitem[{Noh and Han(2016)}]{Noh2016TrainingRA}
Hyeonwoo Noh and Bohyung Han. 2016.
\newblock \href {https://api.semanticscholar.org/CorpusID:2952957} {Training recurrent answering units with joint loss minimization for vqa}.
\newblock \emph{ArXiv}, abs/1606.03647.

\bibitem[{Ouyang et~al.(2022)Ouyang, Wu, Jiang, Almeida, Wainwright, Mishkin, Zhang, Agarwal, Slama, Ray, Schulman, Hilton, Kelton, Miller, Simens, Askell, Welinder, Christiano, Leike, and Lowe}]{Ouyang2022TrainingLM}
Long Ouyang, Jeff Wu, Xu~Jiang, Diogo Almeida, Carroll~L. Wainwright, Pamela Mishkin, Chong Zhang, Sandhini Agarwal, Katarina Slama, Alex Ray, John Schulman, Jacob Hilton, Fraser Kelton, Luke~E. Miller, Maddie Simens, Amanda Askell, Peter Welinder, Paul~Francis Christiano, Jan Leike, and Ryan~J. Lowe. 2022.
\newblock \href {https://api.semanticscholar.org/CorpusID:246426909} {Training language models to follow instructions with human feedback}.
\newblock \emph{ArXiv}, abs/2203.02155.

\bibitem[{Pal et~al.(2022)Pal, Umapathi, and Sankarasubbu}]{Pal2022MedMCQAA}
Ankit Pal, Logesh~Kumar Umapathi, and Malaikannan Sankarasubbu. 2022.
\newblock \href {https://api.semanticscholar.org/CorpusID:247763070} {Medmcqa : A large-scale multi-subject multi-choice dataset for medical domain question answering}.
\newblock In \emph{ACM Conference on Health, Inference, and Learning}.

\bibitem[{Poco and Heer(2017)}]{Poco2017ReverseEngineeringVR}
Jorge Poco and Jeffrey Heer. 2017.
\newblock \href {https://api.semanticscholar.org/CorpusID:7045290} {Reverse‐engineering visualizations: Recovering visual encodings from chart images}.
\newblock \emph{Computer Graphics Forum}, 36.

\bibitem[{Ram et~al.(2021)Ram, Kirstain, Berant, Globerson, and Levy}]{Ram2021FewShotQA}
Ori Ram, Yuval Kirstain, Jonathan Berant, Amir Globerson, and Omer Levy. 2021.
\newblock \href {https://api.semanticscholar.org/CorpusID:230433978} {Few-shot question answering by pretraining span selection}.
\newblock In \emph{Annual Meeting of the Association for Computational Linguistics}.

\bibitem[{Sampat et~al.(2020)Sampat, Yang, and Baral}]{Sampat2020VisuoLingusticQA}
Shailaja~Keyur Sampat, Yezhou Yang, and Chitta Baral. 2020.
\newblock \href {https://api.semanticscholar.org/CorpusID:222225265} {Visuo-lingustic question answering (vlqa) challenge}.
\newblock In \emph{Findings}.

\bibitem[{Susnjak(2022)}]{Susnjak2022ChatGPTTE}
Teo Susnjak. 2022.
\newblock \href {https://api.semanticscholar.org/CorpusID:254853785} {Chatgpt: The end of online exam integrity?}
\newblock \emph{ArXiv}, abs/2212.09292.

\bibitem[{Thirunavukarasu et~al.(2023)Thirunavukarasu, Ting, Elangovan, Gutierrez, Tan, and Ting}]{Thirunavukarasu2023LargeLM}
Arun~James Thirunavukarasu, Darren Shu~Jeng Ting, Kabilan Elangovan, Laura Gutierrez, Ting~Fang Tan, and Daniel Shu~Wei Ting. 2023.
\newblock \href {https://api.semanticscholar.org/CorpusID:259947046} {Large language models in medicine}.
\newblock \emph{Nature Medicine}, 29:1930--1940.

\bibitem[{Thiruvanantharajah et~al.(2021)Thiruvanantharajah, Hangarangoda, and Rajapakshe}]{Thiruvanantharajah2021AutomatedQA}
Maheraj Thiruvanantharajah, Nawanjana Hangarangoda, and S.C. Rajapakshe. 2021.
\newblock \href {https://api.semanticscholar.org/CorpusID:245706713} {Automated question and answer generating system for educational platforms}.
\newblock \emph{2021 6th International Conference on Information Technology Research (ICITR)}, pages 1--6.

\bibitem[{Vo(2024)}]{Vo2024ViMistralXBA}
James Vo. 2024.
\newblock \href {https://api.semanticscholar.org/CorpusID:268681407} {Vi-mistral-x: Building a vietnamese language model with advanced continual pre-training}.
\newblock \emph{ArXiv}, abs/2403.15470.

\bibitem[{Wei et~al.(2022)Wei, Wang, Schuurmans, Bosma, hsin Chi, Xia, Le, and Zhou}]{Wei2022ChainOT}
Jason Wei, Xuezhi Wang, Dale Schuurmans, Maarten Bosma, Ed~Huai hsin Chi, F.~Xia, Quoc Le, and Denny Zhou. 2022.
\newblock \href {https://api.semanticscholar.org/CorpusID:246411621} {Chain of thought prompting elicits reasoning in large language models}.
\newblock \emph{ArXiv}, abs/2201.11903.

\bibitem[{Wu et~al.(2023{\natexlab{a}})Wu, Bansal, Zhang, Wu, Zhang, Zhu, Li, Jiang, Zhang, and Wang}]{Wu2023AutoGenEN}
Qingyun Wu, Gagan Bansal, Jieyu Zhang, Yiran Wu, Shaokun Zhang, Erkang Zhu, Beibin Li, Li~Jiang, Xiaoyun Zhang, and Chi Wang. 2023{\natexlab{a}}.
\newblock \href {https://api.semanticscholar.org/CorpusID:260925901} {Autogen: Enabling next-gen llm applications via multi-agent conversation framework}.
\newblock \emph{ArXiv}, abs/2308.08155.

\bibitem[{Wu et~al.(2023{\natexlab{b}})Wu, Irsoy, Lu, Dabravolski, Dredze, Gehrmann, Kambadur, Rosenberg, and Mann}]{Wu2023BloombergGPTAL}
Shijie Wu, Ozan Irsoy, Steven Lu, Vadim Dabravolski, Mark Dredze, Sebastian Gehrmann, Prabhanjan Kambadur, David Rosenberg, and Gideon Mann. 2023{\natexlab{b}}.
\newblock \href {https://api.semanticscholar.org/CorpusID:257833842} {Bloomberggpt: A large language model for finance}.
\newblock \emph{ArXiv}, abs/2303.17564.

\bibitem[{Wu et~al.(2023{\natexlab{c}})Wu, Jia, Zhang, Li, Zhu, Wang, Lee, Peng, Wu, and Wang}]{Wu2023AnES}
Yiran Wu, Feiran Jia, Shaokun Zhang, Han-Tai Li, Erkang Zhu, Yue Wang, Yin~Tat Lee, Richard Peng, Qingyun Wu, and Chi Wang. 2023{\natexlab{c}}.
\newblock \href {https://api.semanticscholar.org/CorpusID:259063798} {An empirical study on challenging math problem solving with gpt-4}.
\newblock \emph{ArXiv}, abs/2306.01337.

\bibitem[{Xia et~al.(2022)Xia, Chiu, Zhou, Chai, and Cheng}]{Xia2022SystematicLR}
Qi~Xia, Thomas K.~F. Chiu, Xinyan Zhou, Ching~Sing Chai, and Miaoting Cheng. 2022.
\newblock \href {https://api.semanticscholar.org/CorpusID:254969549} {Systematic literature review on opportunities, challenges, and future research recommendations of artificial intelligence in education}.
\newblock \emph{Comput. Educ. Artif. Intell.}, 4:100118.

\bibitem[{Xu et~al.(2015)Xu, Ba, Kiros, Cho, Courville, Salakhutdinov, Zemel, and Bengio}]{Xu2015ShowAA}
Ke~Xu, Jimmy Ba, Ryan Kiros, Kyunghyun Cho, Aaron~C. Courville, Ruslan Salakhutdinov, Richard~S. Zemel, and Yoshua Bengio. 2015.
\newblock \href {https://api.semanticscholar.org/CorpusID:1055111} {Show, attend and tell: Neural image caption generation with visual attention}.
\newblock In \emph{International Conference on Machine Learning}.

\bibitem[{Xu et~al.(2022)Xu, Li, Zhang, Zhou, Bing, Lam, and Si}]{Xu2022FromCT}
Weiwen Xu, Xin Li, Wenxuan Zhang, Meng Zhou, Lidong Bing, Wai Lam, and Luo Si. 2022.
\newblock \href {https://api.semanticscholar.org/CorpusID:254535672} {From clozing to comprehending: Retrofitting pre-trained language model to pre-trained machine reader}.
\newblock \emph{ArXiv}, abs/2212.04755.

\bibitem[{Yan et~al.(2023)Yan, Sha, Zhao, Li, Mart{\'i}nez-Maldonado, Chen, Li, Jin, and Gaevi}]{Yan2023PracticalAE}
Lixiang Yan, Lele Sha, Linxuan Zhao, Yuheng Li, Roberto Mart{\'i}nez-Maldonado, Guanliang Chen, Xinyu Li, Yueqiao Jin, and Dragan Gaevi. 2023.
\newblock \href {https://api.semanticscholar.org/CorpusID:260125084} {Practical and ethical challenges of large language models in education: A systematic scoping review}.
\newblock \emph{Br. J. Educ. Technol.}, 55:90--112.

\bibitem[{Yang et~al.(2023)Yang, Liu, and Wang}]{Yang2023FinGPTOF}
Hongyang Yang, Xiao-Yang Liu, and Chris Wang. 2023.
\newblock \href {https://api.semanticscholar.org/CorpusID:259129734} {Fingpt: Open-source financial large language models}.
\newblock \emph{ArXiv}, abs/2306.06031.

\bibitem[{Yang et~al.(2015)Yang, He, Gao, Deng, and Smola}]{Yang2015StackedAN}
Zichao Yang, Xiaodong He, Jianfeng Gao, Li~Deng, and Alex Smola. 2015.
\newblock \href {https://api.semanticscholar.org/CorpusID:8849206} {Stacked attention networks for image question answering}.
\newblock \emph{2016 IEEE Conference on Computer Vision and Pattern Recognition (CVPR)}, pages 21--29.

\bibitem[{Yuan et~al.(2023)Yuan, Yuan, Tan, Wang, and Huang}]{Yuan2023HowWD}
Zheng Yuan, Hongyi Yuan, Chuanqi Tan, Wei Wang, and Songfang Huang. 2023.
\newblock \href {https://api.semanticscholar.org/CorpusID:257952500} {How well do large language models perform in arithmetic tasks?}
\newblock \emph{ArXiv}, abs/2304.02015.

\bibitem[{Zeng et~al.(2023)Zeng, Gan, Wang, Liu, and Yu}]{Zeng2023LargeLM}
Fanlong Zeng, Wensheng Gan, Yongheng Wang, Ning Liu, and Philip~S. Yu. 2023.
\newblock \href {https://api.semanticscholar.org/CorpusID:265149884} {Large language models for robotics: A survey}.
\newblock \emph{ArXiv}, abs/2311.07226.

\bibitem[{Zhang et~al.(2023)Zhang, Aljunied, Gao, Chia, and Bing}]{Zhang2023M3ExamAM}
Wenxuan Zhang, Sharifah~Mahani Aljunied, Chang Gao, Yew~Ken Chia, and Lidong Bing. 2023.
\newblock \href {https://api.semanticscholar.org/CorpusID:259108959} {M3exam: A multilingual, multimodal, multilevel benchmark for examining large language models}.
\newblock \emph{ArXiv}, abs/2306.05179.

\end{thebibliography}

\appendix

% \section{Example Appendix}
% \label{sec:appendix}

% This is an appendix.

\end{document}